\documentclass[letterpaper, 10 pt, journal, twoside]{ieeetran}
\usepackage{graphicx} 
\usepackage{times} 
\usepackage{amsmath} 
\usepackage{amssymb}  
\usepackage{amsfonts}
\usepackage{array}
\usepackage{algorithm}
\usepackage[noend]{algpseudocode}
\usepackage{multirow,color}
\usepackage{cite}
\usepackage{algpseudocode}
\usepackage{varwidth}
\usepackage{booktabs}
\usepackage{gensymb}
\usepackage[hyphens]{url}
\usepackage[export]{adjustbox}
\usepackage[font=footnotesize]{caption}

\usepackage[inline]{enumitem} 
\usepackage{dblfloatfix}
\usepackage{xcolor}
\usepackage{makecell}
\usepackage{textcomp}
\usepackage{gensymb}
\usepackage{amsmath,lipsum}
\usepackage{mathtools}
\usepackage{url}
\usepackage{verbatim}
\usepackage{kotex, graphicx}
\usepackage{subfigure}
\usepackage[hidelinks]{hyperref}
\usepackage{subcaption}
\usepackage{float}
\usepackage{siunitx}
\usepackage{arydshln}
\usepackage{threeparttable}

\newenvironment{IEEEImpStatement}{%
  \vspace{1.5ex}
  \small
    \textbf{\textit{Note to Practitioners—}}
  \bfseries
}{%
  \par\vspace{1.5ex}
}

\begin{document}

\title{Automated Terminal-to-Housing Assembly System for Flat Ribbon Cable Harness}
%
%
%

\author{Eunkyu~Choi, Joonho~Seo, Seungmin~Lee, and~Seokhwan~Jeong*,~\IEEEmembership{Member,~IEEE,}
\thanks{This research was supported by DAEHA CABLE CO., LTD., the National Research Foundation of Korea(NRF) grant funded by the Korea government(MSIT)(RS-2025-16070605), and the Nano Material Technology Development Program through the NRF funded by Ministry of Science and ICT(RS-2025-25442536) (50\%). \textit{(Eunkyu Choi and Joonho Seo contributed equally to this work) (Corresponding author: Seokhwan Jeong)}}
\thanks{E. Choi, J. Seo, S. Lee and S. Jeong are with Department of Mechanical Engineering, Sogang University, Seoul, South Korea (e-mail: greenwarp@sogang.ac.kr; seojh996@sogang.ac.kr; lnhgf9@sogang.ac.kr; seokhwan@sogang.ac.kr).}%

}

%
%

\markboth{}{} 
%



\maketitle

\begin{center}
\footnotesize
This work has been submitted to the IEEE for possible publication. Copyright may be transferred without notice, after which this version may no longer be accessible.
\end{center}


\begin{abstract}
This paper presents a sensor-minimal automated assembly system for bidirectional single-row flat ribbon cable harnesses (FRCHs). Unlike conventional peg-in-hole or single-terminal insertion tasks, FRCH assembly involves mechanically coupled multi-terminal insertion under flexible and dense geometric constraints. To address this problem, the proposed system performs the assembly through a purely mechanical sequence consisting of Cable Alignment, Lean \& Slide, Weaving, and Clamping, without relying on active sensing or vision. Each mechanism is designed to progressively reduce correlated terminal misalignment, insertion interference, and instability before final locking. Experiments on bidirectional single-row FRCHs achieved an 83.75\% end-to-end process success rate over 80 trials, with success rates of 85.0\% and 82.5\% in the first and second halves, respectively. The cycle time was 33 s under half-speed operation. To the best of our knowledge, this work presents the first automated prototype for FRCH terminal-to-housing assembly for multi-pin housings.
\end{abstract}

\begin{IEEEImpStatement}
Terminal-to-housing assembly for flat ribbon cable harnesses (FRCHs) remains difficult to automate because multiple terminals are mechanically coupled by a flexible ribbon, making independent alignment and insertion infeasible. Conventional automation methods developed for peg-in-hole or discrete wire harness assembly are therefore not directly applicable. This paper presents a sensor-minimal mechanical assembly system that performs FRCH assembly through Cable Alignment, Lean \& Slide, Weaving, and Clamping without relying on vision or active force sensing. For practitioners, the proposed system provides a compact and low-complexity automation concept for repetitive FRCH assembly tasks that still depend heavily on manual labor. Although the current prototype has been validated only at laboratory scale, it demonstrates the feasibility of automated FRCH assembly and provides a foundation for future automation of flexible multi-terminal insertion tasks.
\end{IEEEImpStatement}



%
\IEEEpeerreviewmaketitle

\section{Introduction}
%
%
%
%

\begin{figure}[t]
    \centering
    \includegraphics[width=7.8cm]{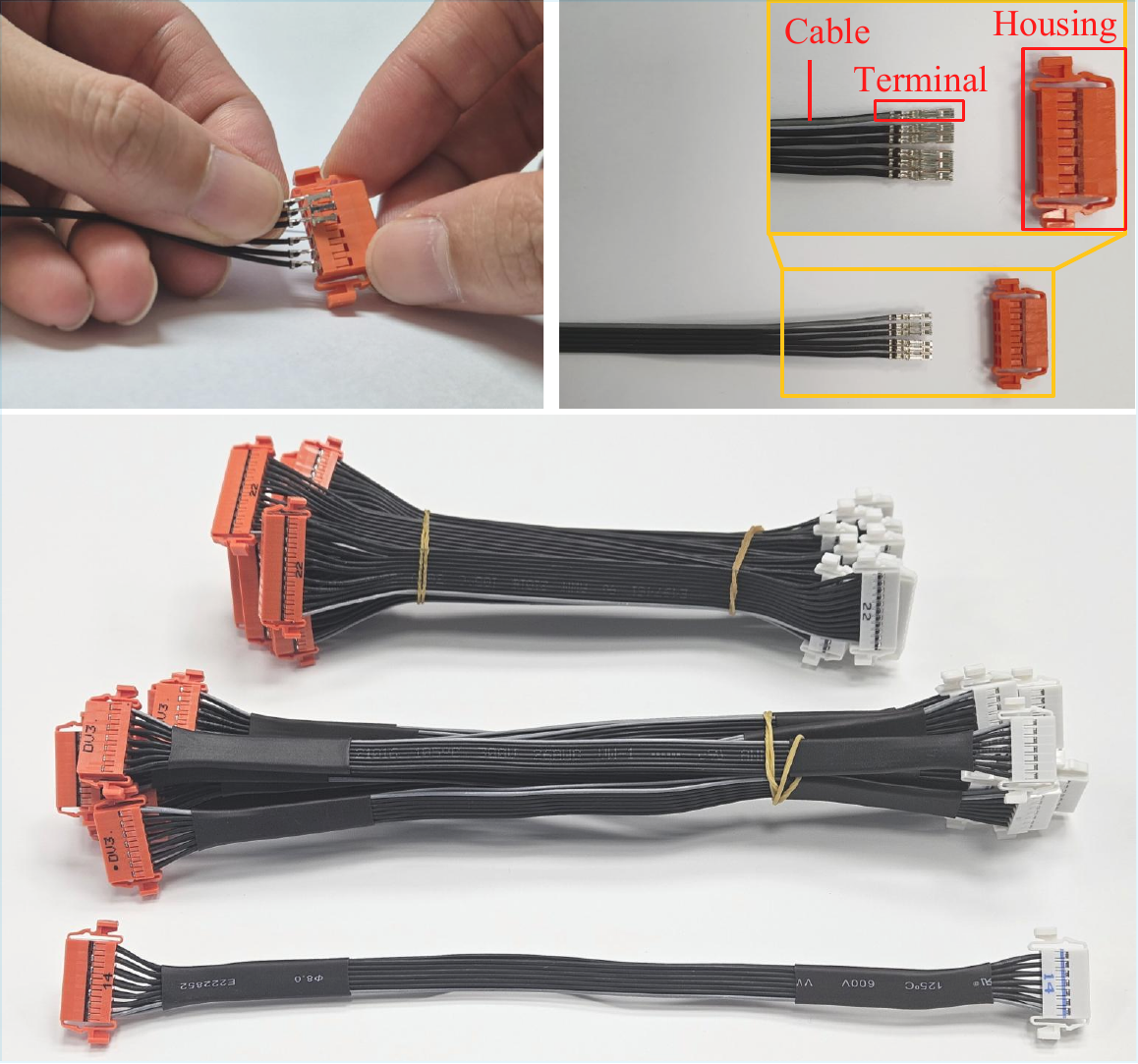}
    \caption{Representative photograph of the wire harness assembly consisting of cables, metallic terminals, and plastic housings. And  examples of assembled harness bundles with different connector configurations.}
    \label{fig:wire harness}
    \vspace{-0.7cm}
\end{figure}

\IEEEPARstart{W}{ire} harnesses are widely used as standard products for the efficient transmission of power and electrical signals across diverse industries, including electronics, automotive, and robotics\cite{karlsson2024automatic, hernandezmejia2026review, navasreascos2026simulation}. They consist of multiple flexible cables, metal terminals attached to the cable ends, and plastic housings that align and accommodate these terminals (see Fig.~\ref{fig:wire harness}); the assembled structure is referred to as a \textit{wire harness assembly}. Several upstream manufacturing processes, such as wire stripping, cutting, and terminal crimping, have already been widely automated. However, the terminal-housing assembly step remains a major bottleneck in wire harness production\cite{lorenz2023approaches} and this limitation becomes more severe when multiple cable terminals must be inserted into a housing under tight geometric constraints.

\par 

\begin{figure*}[t]
    \centering
    \subfigure[]{
        \includegraphics[width=0.3\textwidth]{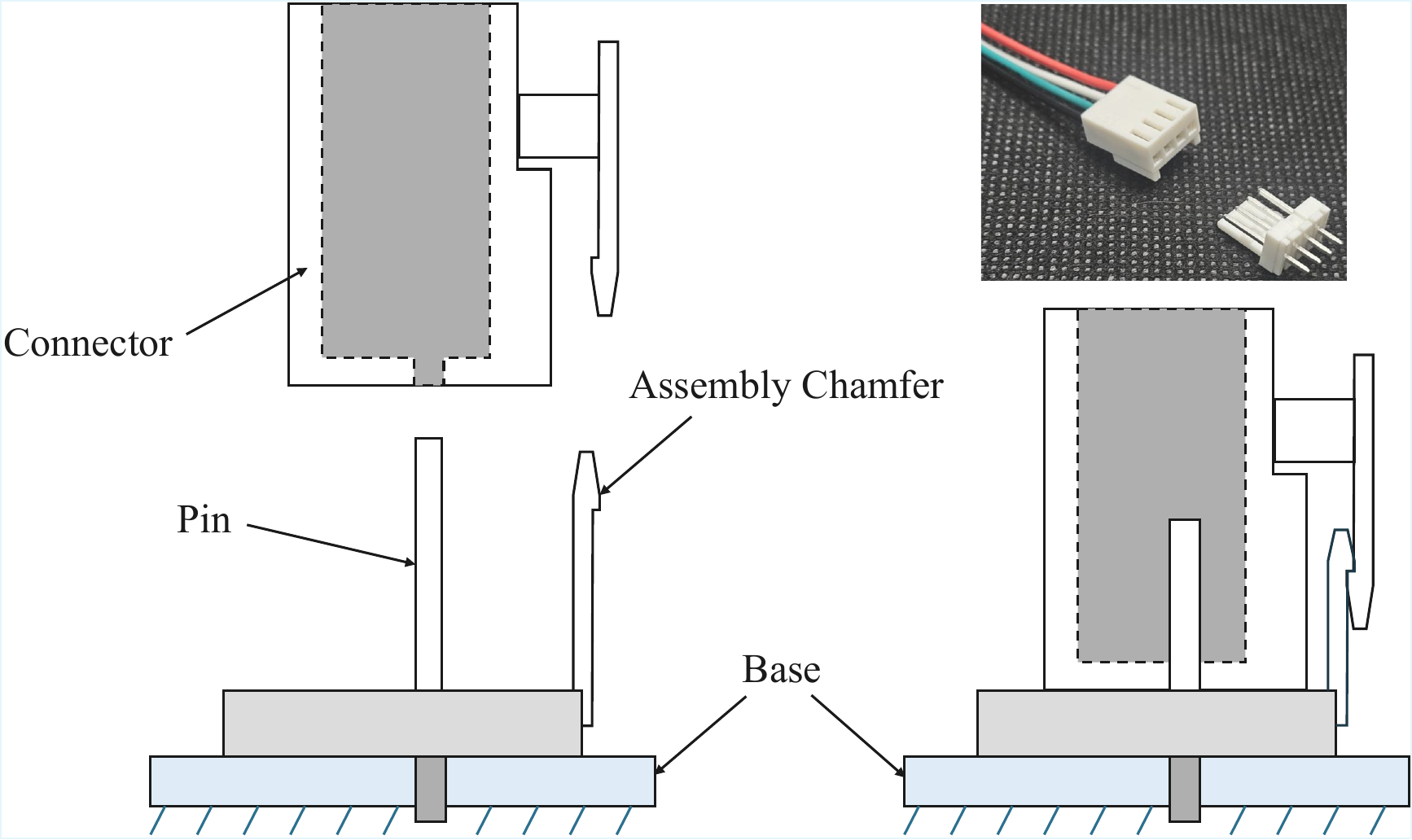}
        \label{fig:connector header}
     }
    \hfill
    \subfigure[]{
        \includegraphics[width=0.3\textwidth]{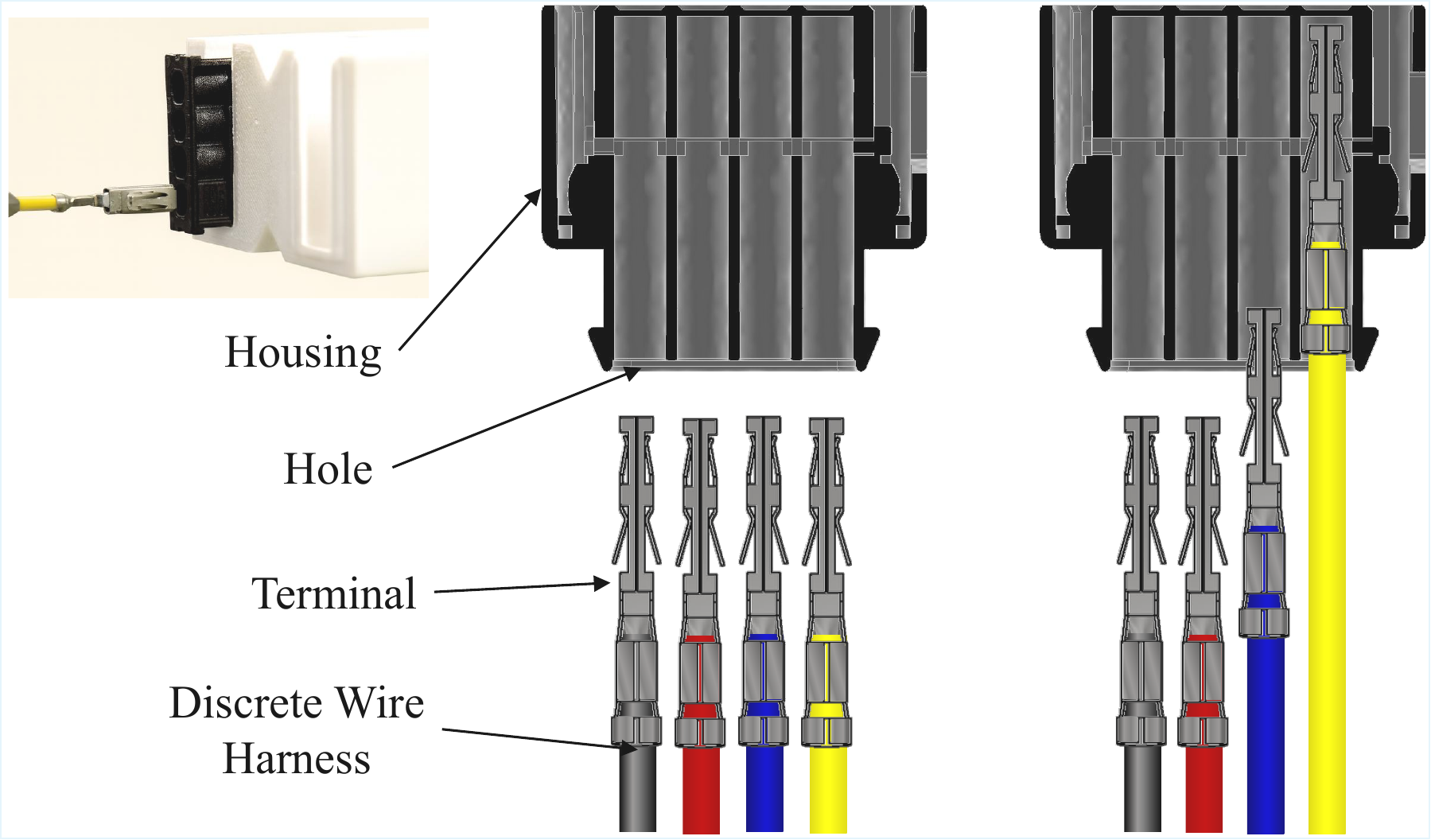}
        \label{fig:DWH_fig}
    }
    \hfill
    \subfigure[]{
        \includegraphics[width=0.3\textwidth]{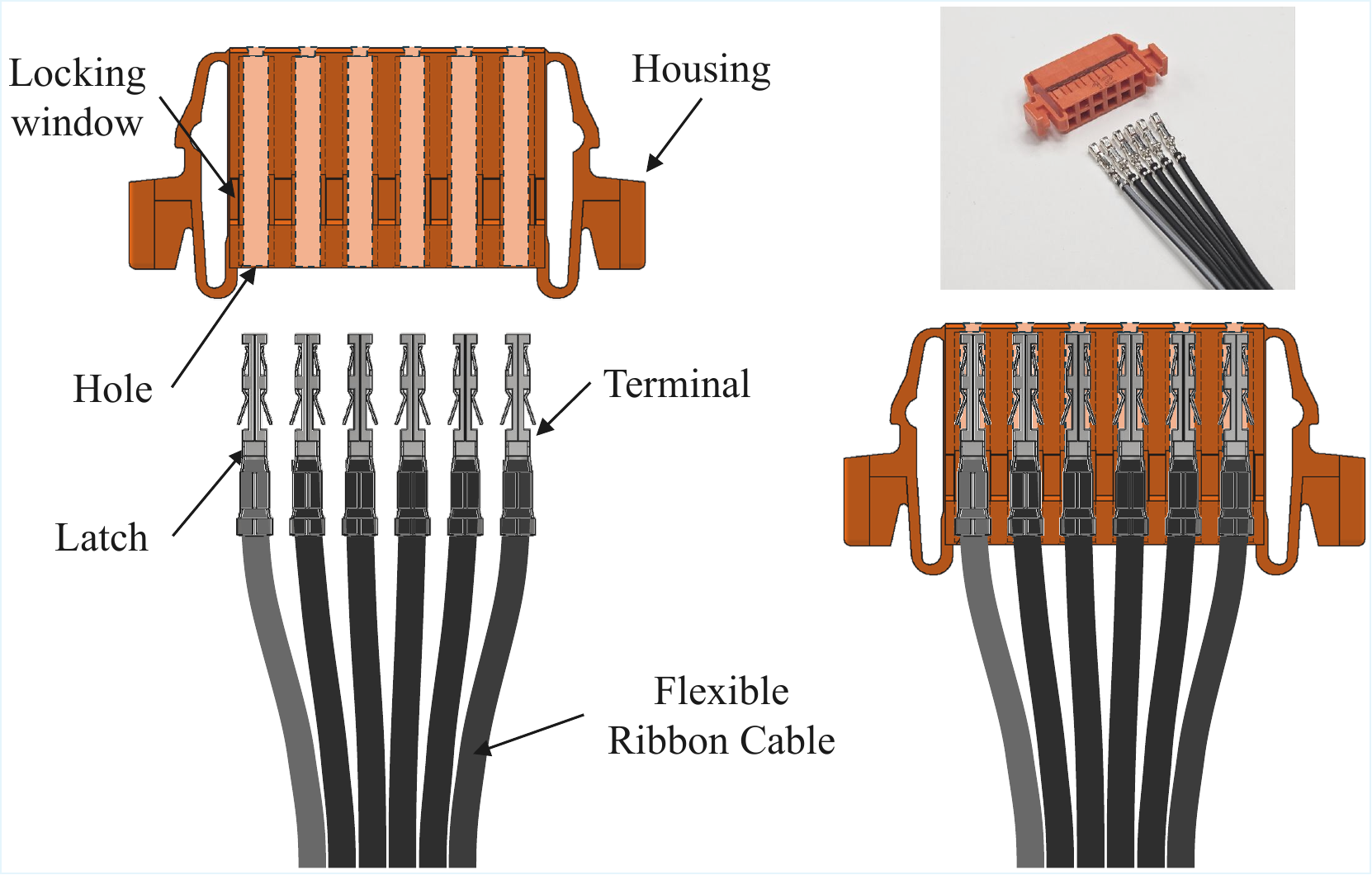}
        \label{fig:FRCH_fig}
    }
    \caption{Comparison between (a) conventional pin-to-connector insertion studied in previous works, (b) terminal-to-housing insertion for discrete wire harness assembly, where individual wires are independently, and (c) the present study focusing on terminal-to-housing insertion for flat ribbon cable harness assembly.}
    \label{fig:previous research}
    \vspace{-0.2cm}
\end{figure*}

\par A representative automation task related to cable assembly is conventional pin-to-connector insertion, as shown in Fig.~\ref{fig:connector header}. Prior studies have largely addressed such tasks as a peg-in-hole problem in applications involving harness-to-board or cable detection.\cite{NGUYEN2024360,9268291,ying2022pose}. 
In force/torque-guided assembly, Mei et al.\cite{10746551}, Xing et al.\cite{9064951}, and Zhang et al.\cite{8365152} improved insertion robustness in peg-in-hole tasks through wrist force sensing, passive compliance with force feedback, and jamming-aware force control, respectively. In vision-guided assembly, Xu et al.\cite{9748081}, Nguyen and Yoon\cite{NGUYEN2021365} proposed vision-based methods for peg-in-hole alignment and wire-harness profile extraction, respectively. Mou et al.\cite{MOU2022105164} and Chen et al.\cite{9650911} further combined visual perception with force-aware insertion strategies, such as admittance control and hybrid force/position control, to handle contact-rich or deformable insertion tasks. De Gregorio et al.\cite{8395267} integrated robotic vision and tactile sensing for wire-terminal insertion. More recently, learning-based approaches have also been introduced: Men et al.\cite{10220113} proposed a policy-fusion transfer framework for peg-in-hole insertion, Hao and Xu\cite{10721256} addressed relay-socket assembly as a multiple peg-in-hole problem using an automated control method, and Beck et al.\cite{beck2025deep} showed that deep-learning-based socket localization can improve automated electrical connector mating. 
\par Although these studies demonstrated strong performance, most of them focused on rigid or locally correctable insertion tasks, such as peg-in-hole assembly, single cable-connector insertion, wire-terminal insertion, or socket localization. Their applicability remains limited when multiple densely arranged terminals must be inserted simultaneously under flexible and mechanically coupled conditions.

 \par A more relevant point of comparison is terminal-to-housing insertion in discrete wire harnesses (DWHs), shown in Fig.~\ref{fig:DWH_fig}. In DWHs, each cable is physically separated, allowing each terminal to be individually manipulated, aligned, and inserted. This structural independence has made single-terminal insertion a practical automation strategy and has supported its industrial adoption. In contrast, flat ribbon cable harnesses (FRCHs), shown in Fig. ~\ref{fig:FRCH_fig}, mechanically couple multiple terminals through a shared flexible ribbon. Consequently, the terminals cannot be corrected independently during insertion, and assembly must proceed under dense spatial and geometric constraints. The problem therefore shifts from independent single-terminal insertion to ribbon-coupled multi-terminal insertion. For this reason, the automation paradigm established for DWHs does not directly extend to FRCHs, whose terminal-to-housing assembly still remains largely dependent on manual labor in practice\cite{hernandezmejia2026review}.
\par  This structural coupling gives rise to three major challenges. First, terminal pose errors can no longer be corrected independently, because the motion of one terminal propagates to adjacent terminals through the shared ribbon. Cable compliance further amplifies this uncertainty during handling and insertion. Second, the narrow terminal pitch (2.0mm-2.5mm) and shared ribbon geometry make the process highly sensitive to small pose deviations and increase the likelihood of mechanical interference among adjacent terminals during insertion. Third, the partially inserted state is difficult to stabilize, because contact with internal housing features may cause catching, rebound, or loss of alignment before final locking. FRCH terminal-to-housing assembly is therefore not a straightforward and incremental extension of peg-in-hole or single-terminal insertion, but a distinct automation problem of FRCH assembly.

\par To the best of our knowledge, this work presents the first reported automated prototype system for FRCH terminal-to-housing assembly. Unlike conventional automation approaches developed for peg-in-hole or independent single-terminal insertion, the proposed system is specifically designed to address the mechanically coupled, interference-prone, and insertion-sensitive nature of FRCHs. Instead of relying on complex sensing or dexterous manipulation, it adopts a sensor-minimal and mechanically structured strategy that progressively resolves the coupled insertion process. Specifically, correlated terminal pose uncertainty is handled through structured guidance and error-tolerant alignment, insertion interference is mitigated through controlled relative motion during engagement, and instability near final locking is reduced by stabilizing the cable-terminal set before full insertion. In this way, the proposed system converts a highly coupled multi-terminal insertion problem into a sequentially manageable assembly process.
\par Experiments on bidirectional single-row FRCHs demonstrate improved pre-insertion performance and reliable full-process assembly, thereby supporting the feasibility of the proposed design for automated FRCH terminal-to-housing assembly. The main contributions of this paper are as follows:
\begin{enumerate} 

\item A problem formulation that characterizes FRCH terminal-to-housing assembly. 
\item A purely mechanical, sensor-minimal assembly strategy that addresses correlated terminal misalignment, insertion interference, and instability during final engagement.
\item An automated prototype system for FRCH terminal-to-housing assembly.

\end{enumerate}

\par This paper is organized as follows with a \textit{Supplementary Video}: Section \ref{sec:Challenge} describes the mechanism and modeling, design of each assembly procedure for the proposed FRCH assembly system. The performance of the prototype of the proposed system is validated by assembly experiments in Section \ref{sec:exp}. The discussion and conclusion are followed in Section \ref{sec:dis and con}.
\begin{figure}[t]
    \centering
    \subfigure[]{
    \centering
    \includegraphics[width=0.45\linewidth]{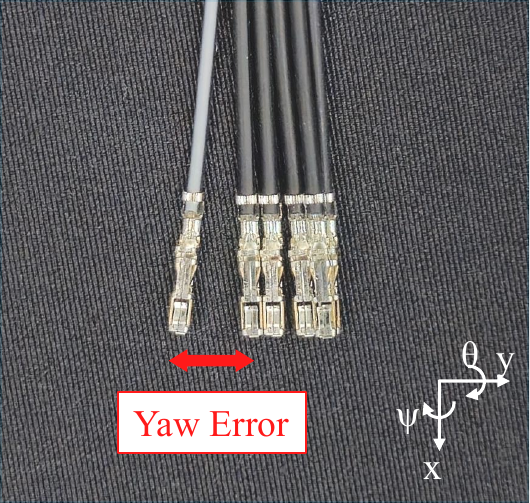}
    \label{fig:yaw error}
    }
    \hfill
    \subfigure[]{
    \centering
    \includegraphics[width=0.45\linewidth]{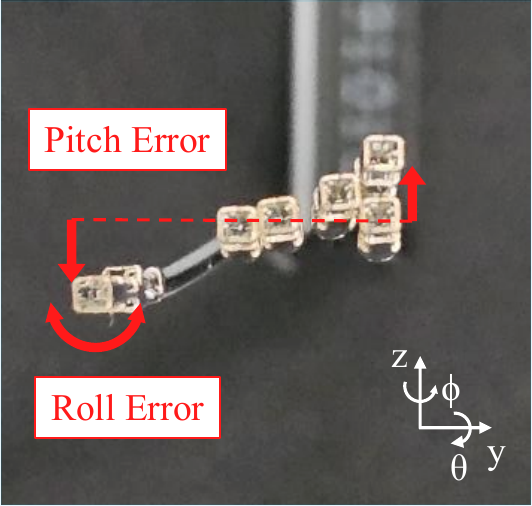}
    \label{fig:pitch error}
    }
    \subfigure[]{
    \centering
    \includegraphics[width=0.95\linewidth]{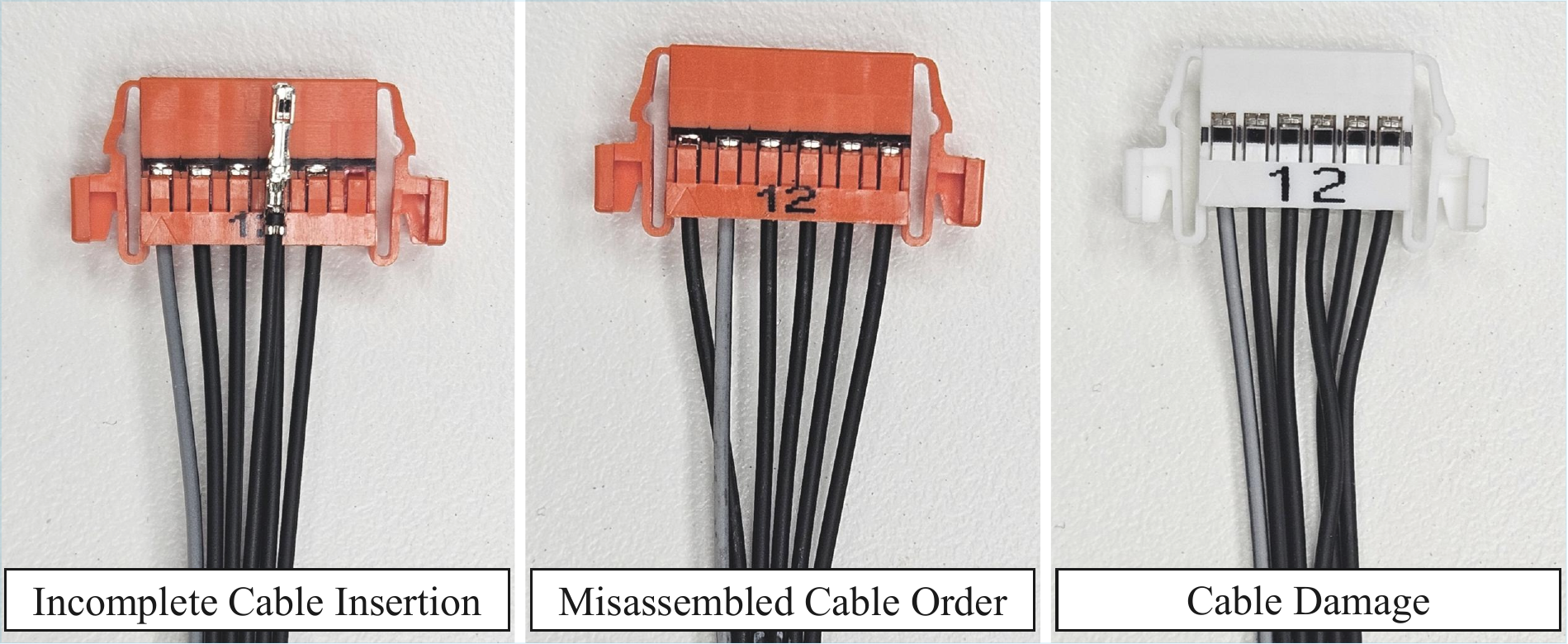}
    \label{fig:pitch error}
    }
    \caption{ Misalignment of harness cable terminals: (a) top view highlighting the lateral offset caused by yaw error, and (b) front view showing angular deviation of individual terminals caused by pitch error. (c) Examples of manual assembly failures in the FRCH process including incomplete cable insertion at the center position, misassembled cable order due to wiring sequence error, and cable damage caused by excessive insertion force.}
    \label{fig:alignment error}
    \vspace{-0.2cm}
\end{figure}
\section{Challenges in Flat Ribbon Cable Harness Assembly}
\label{sec:Challenge}
\par Existing terminal–housing automation systems for DWH can grip each terminal individually and control its position and orientation, which enables active compensation of residual terminal–housing misalignment\cite{8395267,Caporali2026VisionTactile}. In contrast, FRCH does not permit independent manipulation of individual terminals and poses distinct challenges due to inter-cable interference, cable flexibility, terminal pose uncertainty, and the structural complexity of terminals and connector \cite{hernandezmejia2026review,Wang2024ComputerVisionWireHarness,Cho2025CableWiring}. This section introduces the key technical challenges that must be addressed to achieve reliability, repeatability, and high-throughput assembly for FRCH.

\subsection{Terminal Positional Misalignment}
\label{subsec:2A}

\par Terminals in an FRCH bundle remain physically suspended in free space, inducing large pose uncertainty during assembly (see Fig. \ref{fig:alignment error}). In particular, spacing imbalance and directional deviation frequently occur during handling the cable bundle, preventing consistent arrangement\cite{hernandezmejia2026review,Wang2024ComputerVisionWireHarness,Cho2025CableWiring}; such uncertainty can cause interference at housing entrances or collisions between adjacent terminals during insertion, leading to assembly failure\cite{10746551,9064951,8365152,10721256,Hartisch2024ConnectorAssembly}. Precise pre-alignment of terminals to the housing slot pitch is therefore essential to ensure insertion stability. In the bidirectional FRCH (UL 21016 \#26 6c) used in this study, the terminals in a floating state exhibited random orientation errors in roll, pitch, and yaw. In some cases, the measured orientation errors exceeded the specified tolerance range ($\pm 11.58^\circ$, $\pm 3.22^\circ$, and $+ 1.09^\circ$/$- 5.14^\circ$ on the left side, $\pm 11.58^\circ$, $\pm 4.62^\circ$, and $+ 1.56^\circ$/$- 7.24^\circ$ on the right side, respectively). Limiting each error reduces the need for subsequent compensation and establishes the basis for stable operation of the entire mechanism. Pitch deviation remains a primary failure factor. As shown in Fig. \ref{fig:pitch error}, some terminals exhibited pitch deviations exceeding $\pm 10^\circ$, which hindered conformity between the terminal tip and the housing hole and could induce catching or rebound-induced retraction during insertion. In multi-pin configurations, even a slight height deviation in a single terminal can sharply reduce the overall insertion success rate; therefore, pitch error is difficult to resolve by simple pre-alignment and requires continuous correction throughout the insertion process\cite{8365152,10721256}.

\subsection{Insertion Failures and Stability Issues} \label{subsec:2B}

\par Problems arising during terminal–housing assembly can be grouped into two categories. The first involves mechanical interference and incomplete insertion, and the second concerns the pose instability that persists up to the final locking step. This section describes each problem and motivates the corresponding mechanisms.
\begin{figure*}[t]
    \centering
    \includegraphics[width=1\textwidth]{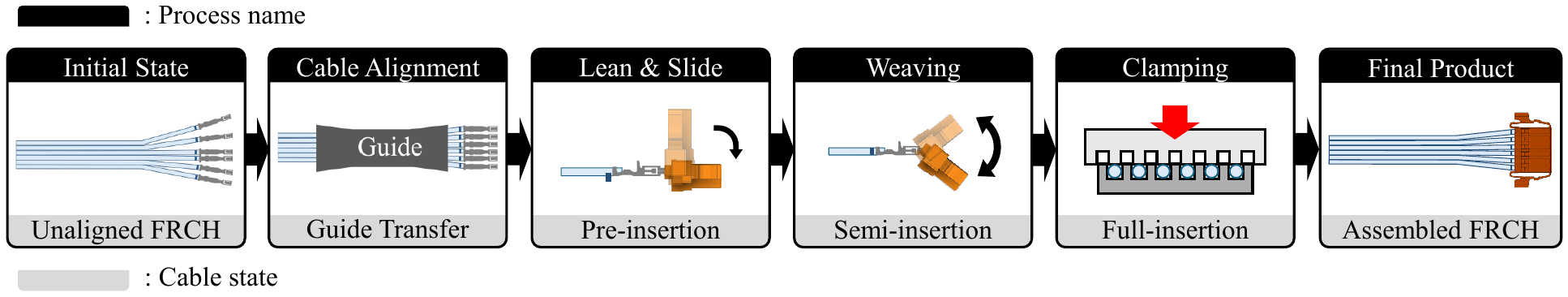}
    \caption{Overall operational workflow of the proposed FRCH assembly process. 
The process begins with Cable Alignment using a guide structure to separate and arrange terminals, followed by Lean \& Slide for pre-insertion and orientation correction, Weaving for semi-insertion with oscillatory motion, and Clamping for full insertion and stabilization, resulting in the final assembled FRCH.}
    \label{fig:Overall_workflow}
    \vspace{-0.2cm}
\end{figure*}

\begin{figure*}[t]
    \subfigure[]{\includegraphics[width=0.49\textwidth, height=5.5cm]{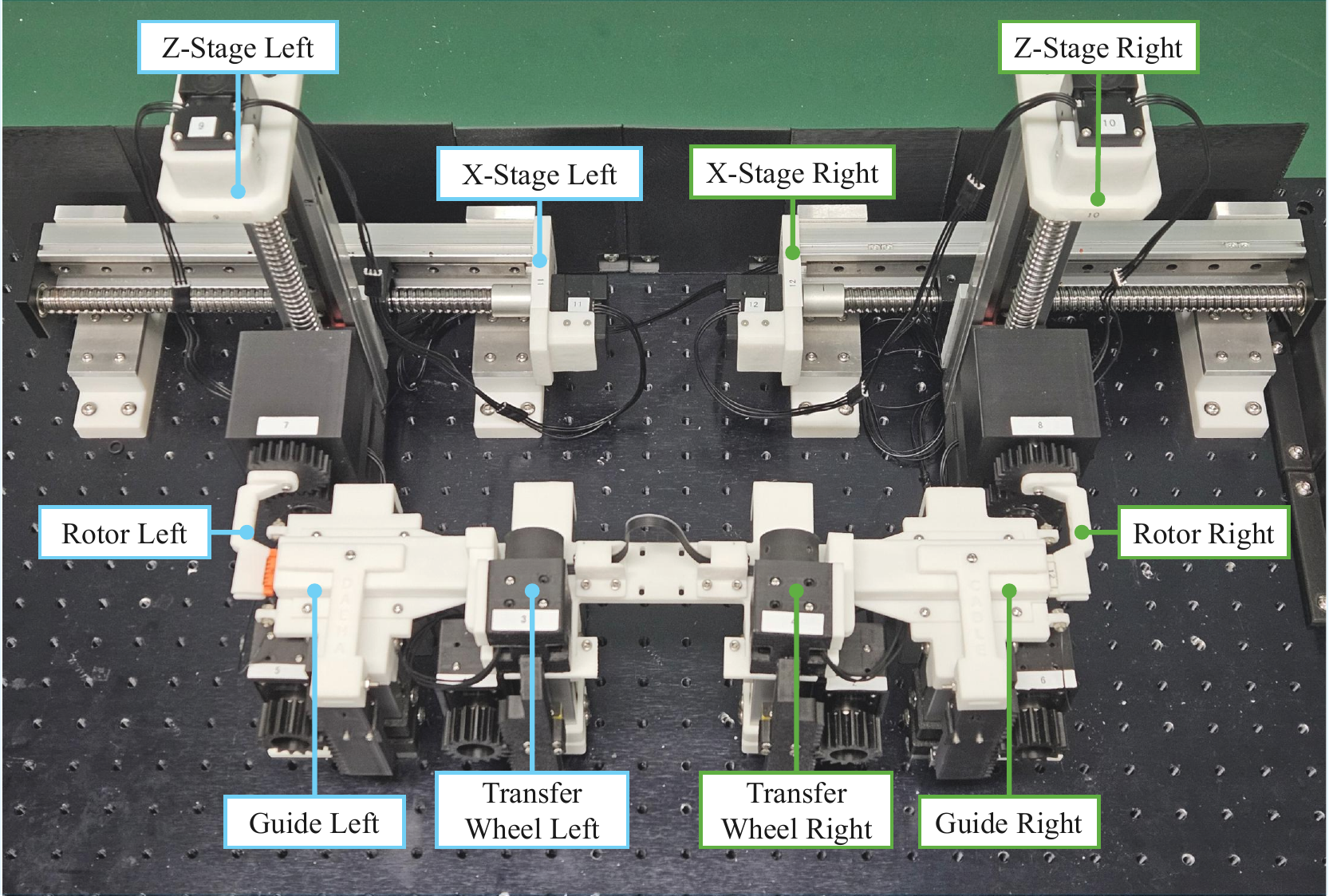}
    \label{fig:system_overview}
    }
    \hfill
    \subfigure[]{
    \centering
    \includegraphics[width=0.48\textwidth, height=5.5cm]{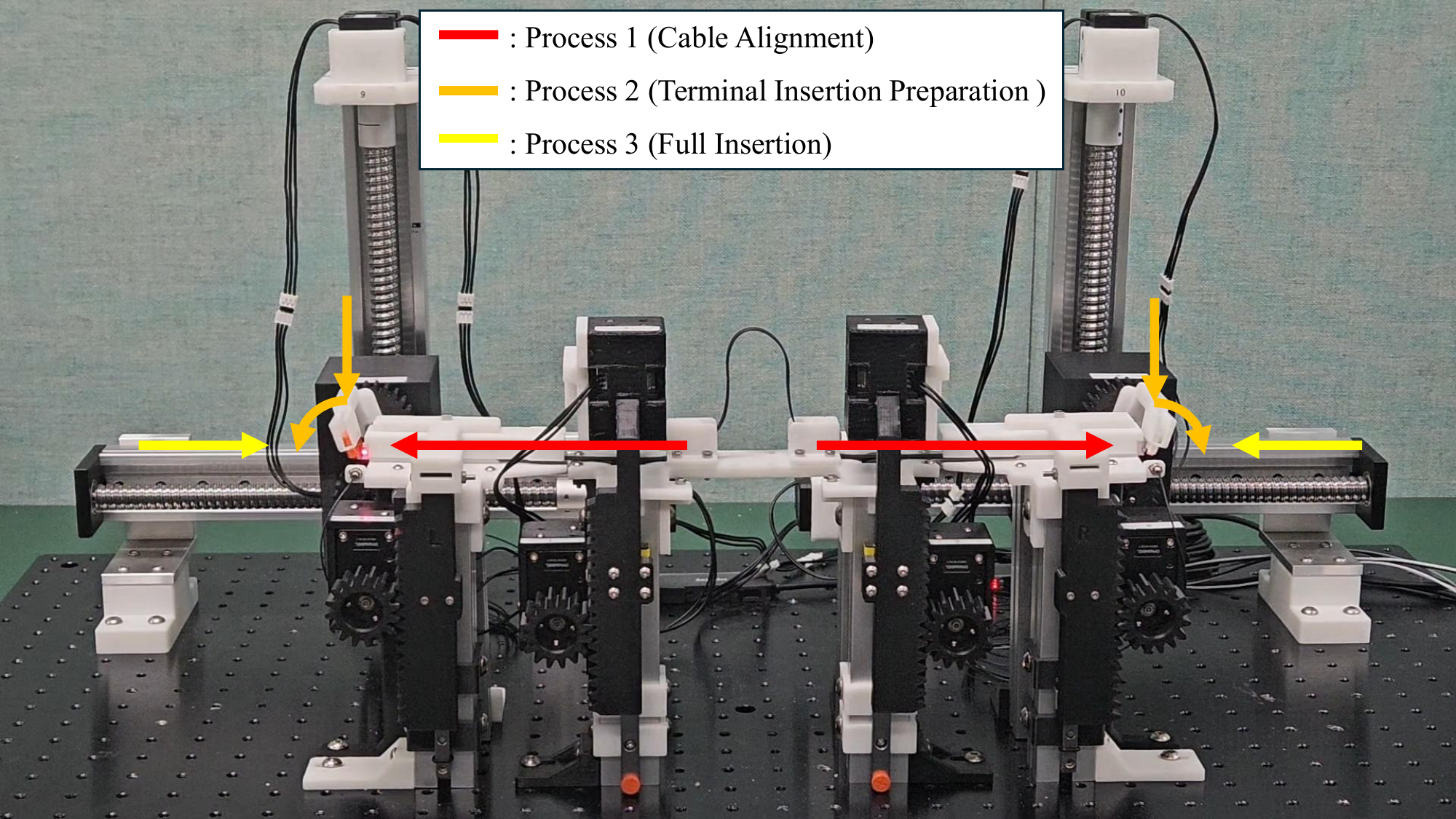}
    \label{fig:process_overview}
    }
    \caption{Configuration of the proposed automated system: (a) (Top view) identification of the main system components, and (b) (Front view) illustration of process steps and corresponding motions of the devices: Process 1 - Cable Alignment, Process 2 - Terminal Insertion using Lean \& Slide, and Process 3 - Full Insertion}
    \label{fig:overall_system}
    \vspace{-0.2cm}
\end{figure*}
\subsubsection{Jamming and Insertion Reliability}
\par Even if terminal–housing alignment is achieved, internal structural features of the housing, such as the locking window (see Fig. \ref{fig:FRCH_fig}), can introduce irregular mechanical interference and resistance during insertion, causing assembly failure\cite{10746551,9064951,8365152,8395267,Hartisch2024ConnectorAssembly}. This phenomenon also appears in DWH assembly; however, when multiple terminals are inserted simultaneously, as in FRCH assembly, the failure of even a subset can propagate to adjacent terminals, dramatically increasing the probability of overall insertion failure.
\par Furthermore, if any single terminal does not reach the specified insertion depth, the likelihood of proceeding to the final locking step drops sharply. A dynamic insertion strategy is required to overcome friction and interference during insertion, absorb residual errors, and advance terminals stably.

\subsubsection{Stability Prior to Final Locking}
\par After terminals enter the housing, terminal poses remain unstable until the final locking step. Rebound forces generated during insertion can induce subtle pose changes that disrupt established terminal alignment and lead to locking failure. In multi-pin configurations, the pose change of one terminal can propagate to adjacent terminals.
\par Full engagement requires an insertion force above a certain threshold\cite{8395267,Hartisch2024ConnectorAssembly}. If the terminal and cable poses are not mechanically constrained, the insertion rebound force can push the entire terminal set backward and cause assembly failure. Mechanical fixation that maintains terminal poses against rebound and external disturbances is essential during the latter phase of insertion.

\section{Flat Ribbon Cable Harness Assembly Process} \label{sec:Mechanism}

\par Fig.~\ref{fig:Overall_workflow} illustrates the overall operational workflow of the proposed FRCH assembly process. 
The process starts from an unaligned FRCH and proceeds through four mechanism-level operations: Cable Alignment, Lean \& Slide, Weaving, and Clamping, resulting in the final assembled FRCH. 
This workflow provides a high-level view of how the proposed mechanisms sequentially perform \textit{Cable Alignment}, \textit{Pre-insertion}, \textit{Semi-insertion}, and \textit{Full insertion}.
\par The overall architecture of the proposed FRCH assembly automation system is shown in Fig.~\ref{fig:overall_system}. 
The system adopts a left-right symmetric configuration for bidirectional FRCH assembly, as shown in Fig.~\ref{fig:system_overview}. 
The automated assembly procedure is organized into three process groups, as shown in Fig.~\ref{fig:process_overview}: 
Process~1 performs cable feeding, transfer, and alignment (\textit{Cable Alignment}); 
Process~2 executes terminal--housing insertion from \textit{Pre-insertion} to \textit{Semi-insertion}; 
and Process~3 completes \textit{Full-insertion}. 
The mechanism-level operations in Fig.~\ref{fig:Overall_workflow} correspond to these processes as follows: Cable Alignment corresponds to Process~1, Lean \& Slide and Weaving constitute Process~2, and Clamping corresponds to Process~3. 
The following subsections describe each process in detail.

\par We target single-row, two-ended automatic assembly of an FRCH (UL 21016 \#26 6c) and the two housings connected to its ends (SMH250-H12S4 / SMH200-H12S4B). The automation system uses twelve servomotors (XM430-W350-T, Robotis) under current-based position control. Terminal presence is detected by an optical fiber sensor (BF5R-D1-N, Autonics). All devices are integrated on an ARM Cortex-M7–based microprocessor board (OpenCR~1.0, Robotis) and controlled by embedded code. We assume that an external interface (e.g., a robot manipulator) feeds the cable into the auto-assembly device as the initial state.

\begin{figure}[t]
    \centering
    \includegraphics[width=7.5cm]{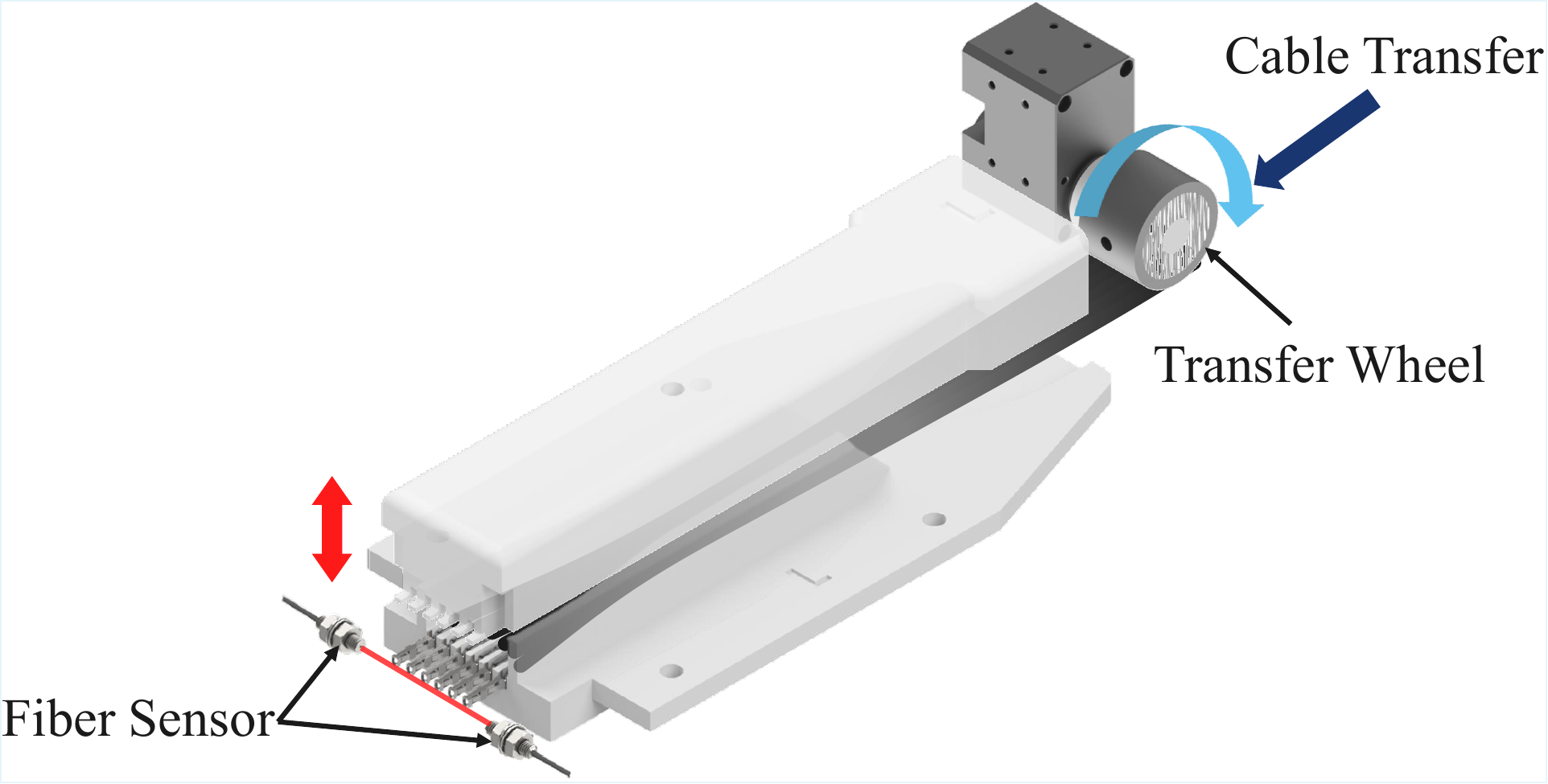}
    \caption{Assembled view of the integrated guide mechanism. The upper and lower guide components move vertically toward each other (red arrow) to adjust the gap, while the cable bundle is inserted horizontally through the central pathway (blue arrow) for alignment and clamping.}
    \label{fig:Aligner Assembled}
    \vspace{-0.2cm}
\end{figure}

\subsection{Process 1 : Cable Alignment (Guide Transfer)} \label{process1_Guide_transfer}

\par This work carries out the assembly of a flat ribbon cable harness with six terminals on each end into two housings: a 6-pin housing with a 2.5mm pitch (SMH250-H12S4) and a 6-pin housing with a 2.0mm pitch (SMH200-H12S4B). As described in Sec.~\ref{subsec:2A}, each compliant terminal in a floating state allows three rotational degrees of freedom—roll, pitch, and yaw—so the two sets of six terminals together yield a total of 36 passive degrees of freedom. This high number of degrees of freedom introduces pose uncertainty during assembly. We therefore propose a semi-enclosed guide to suppress these errors (see Fig.~\ref{fig:Aligner Assembled}).

\par
\begin{figure}[t]
    \centering
    \subfigure[]{
    \centering
    \includegraphics[width=0.29\linewidth, height=3.9cm]{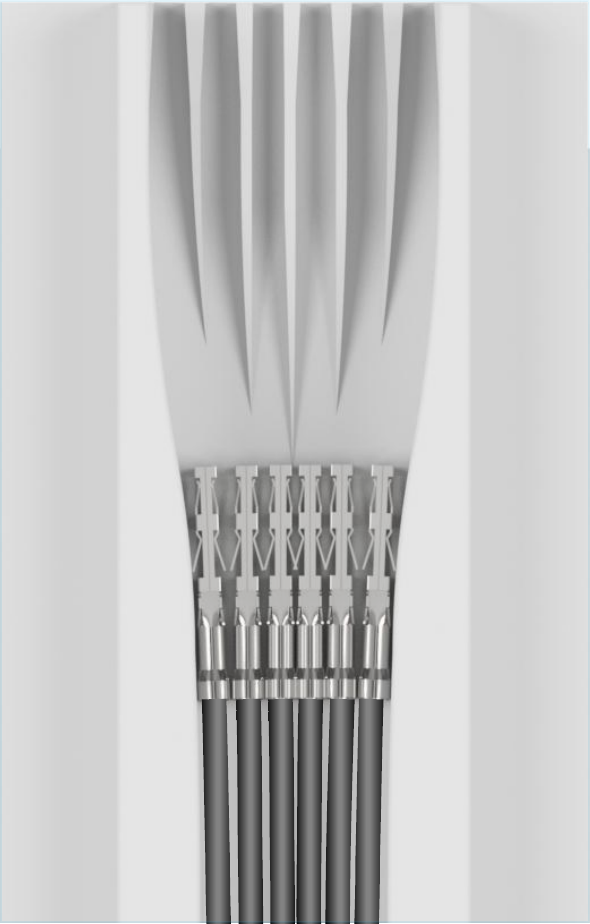}
    \label{fig:top_view3}
    }
    \hfill
    \subfigure[]{
    \centering
    \includegraphics[width=0.29\linewidth]{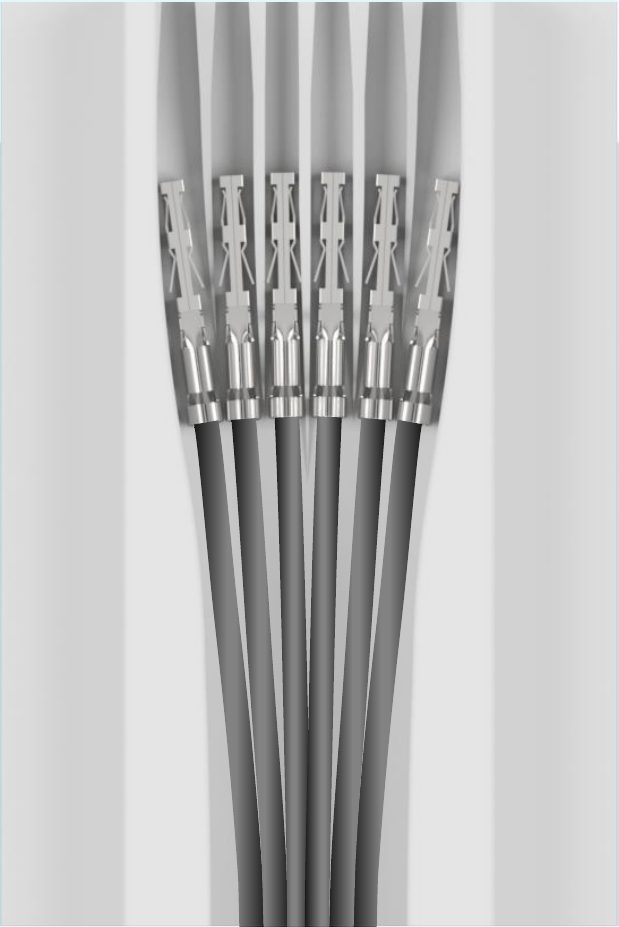}
    \label{fig:top_view2}
    }
    \hfill
    \subfigure[]{
    \centering
    \includegraphics[width=0.29\linewidth]{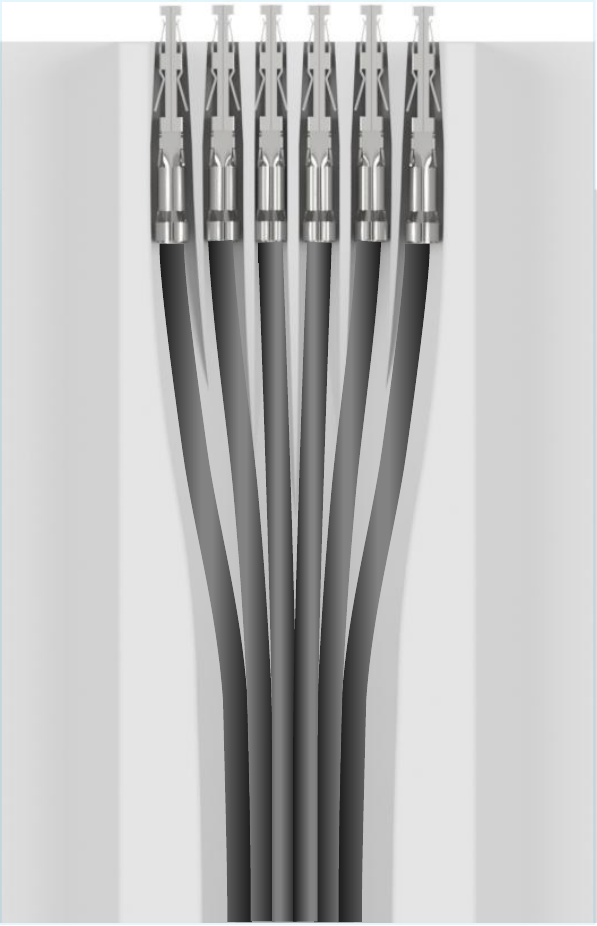}
    \label{fig:top_view1}
    }
    \caption{Sequential process of cables passing through the guide: (a) initial state before contacting the curved surface, (b) intermediate stage of separation by the surface, and (c) final state after complete alignment.}
    \label{fig:Guide_cable transfer}
    \vspace{-0.2cm}
\end{figure}

 The guide consists of a \textit{Guide Top} and a \textit{Guide Bottom}. The Guide Bottom forms the passage through which the terminal-carrying cable travels, and a transfer wheel with a rubber-coated surface feeds the cable into the guide. At this stage, each terminal is passively aligned by the guide to match the housing-slot pitch (see Fig.~\ref{fig:Guide_cable transfer}). The left and right terminal sets are separated to match the pitches of their respective housings, and optical fiber sensors are placed at fixed distances from the guide outlet (left: 5.0mm; right: 3.5mm) to detect in real time whether each terminal tip has reached the assembly standby position. Once a sensor is triggered, cable transfer stops immediately to provide a consistent ready state for assembly. The upper structure functions as a cover to prevent cable escape and subsequently serves as a support that constrains terminal poses during insertion. These steps proceed sequentially at both ends.
\begin{figure}[t]
    \centering
    \subfigure[]{
        \includegraphics[width=0.4\textwidth]{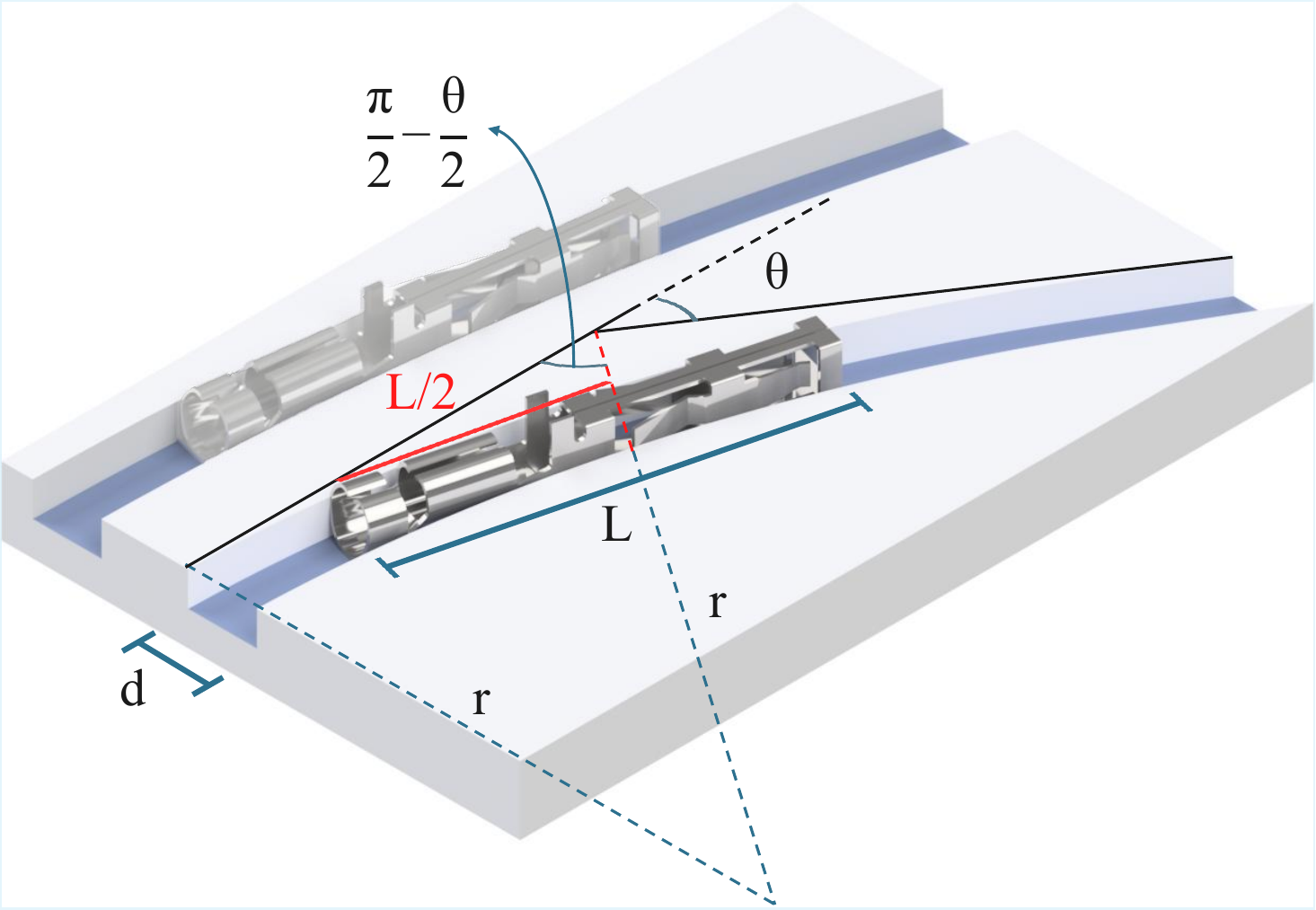}
        \label{fig:curvature of guide}
     }
    \hfill
    \subfigure[]{
        \includegraphics[width=0.4\textwidth]{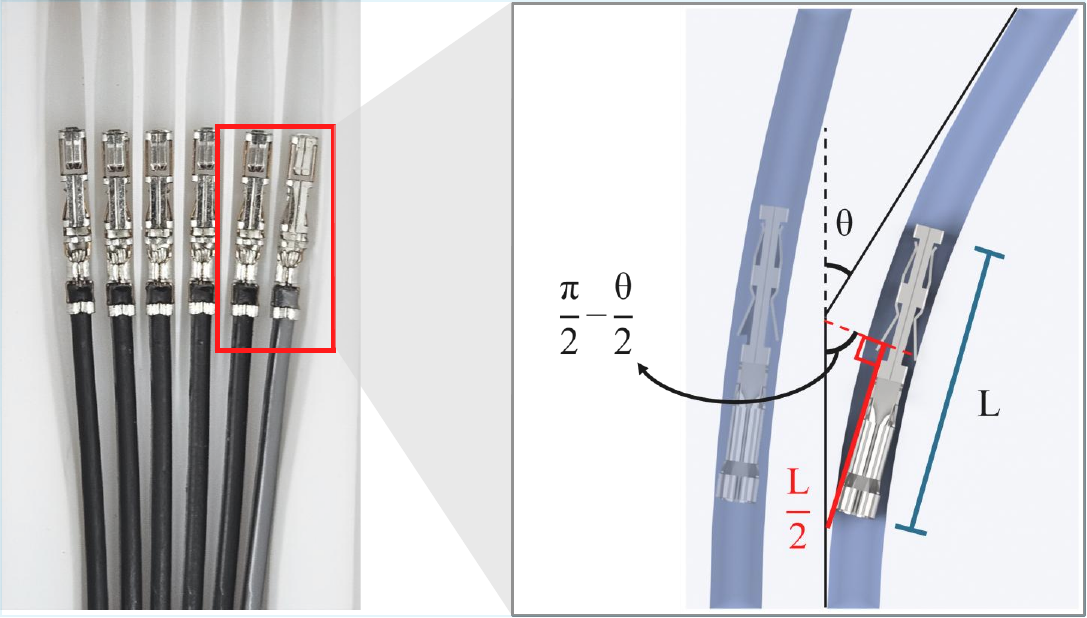}
        \label{fig:curvature enlarge}
    }
    \caption{Guide path geometry used for terminal separation in Process 1: (a) idealized geometric model used to derive the lower-bound fillet radius $r_{min}$ for a rigid terinal of length $L$ traversing a branched guide path with width $d$ and angle $\theta$;(b) actual terminal arrangement and enlarged view of terminal separation in the guide, illustrating the llimited free-length region near the cable end.}
    \label{fig:curvature}
    \vspace{-0.6cm}
\end{figure}
\par The path geometry of the Guide Bottom is designed based on the physical characteristics of the terminals. Unlike the flexible cable body, the crimped metal terminals exhibit relatively high stiffness over the short free length near the cable end. Therefore, the terminal is approximated as a rigid rectangular element when determining the guide-path curvature.

The guide-path model is used as a first-order geometric design guideline rather than as a complete contact-mechanics model. Its purpose is to determine a simple and manufacturable curvature radius that allows the terminals to branch smoothly while reducing geometirc interference. As shown in Fig.~\ref{fig:curvature of guide}, the geometric relationship used to dtermine the guide curvature is expressed as 

\begin{equation}
    r = \frac{\frac{L}{2} \cos\left(\frac{\pi}{2} - \frac{\theta}{2}\right)}{1 - \sin\left(\frac{\pi}{2} - \frac{\theta}{2}\right)} - d
    \label{curvature equation}
\end{equation}
where $L$ denotes the terminal length, $\theta$ denotes the bend angle of the guide path, and $d$ denotes the path width. 

Because the objective of this guide is to provide robust passive alignment through a simple mechanical structure, material deformation, friction, and fit tolerances are not explicitly modeled in Equation \eqref{curvature equation} . Instead, these nonideal effects are handled at the mechanism level in the subsequent assembly sequence: Lean \& Slide reduces residual pitch and yaw misalignment, Weaving alleviates internal catching during insertion, and Clamping suppresses rebound and backward slip during full insertion.
\begin{figure}[t]
    \centering
    \subfigure[]{
    \centering
    \includegraphics[width=0.26\linewidth]{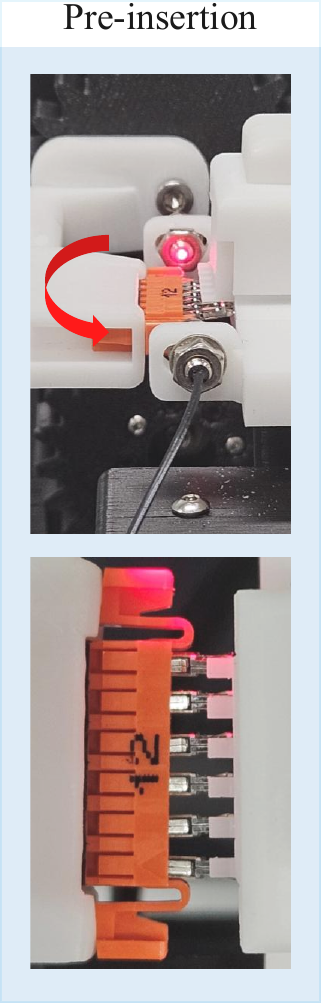}
    \label{pre-insertion}
    }
    \hfill
    \subfigure[]{
    \centering
    \includegraphics[width=0.26\linewidth]{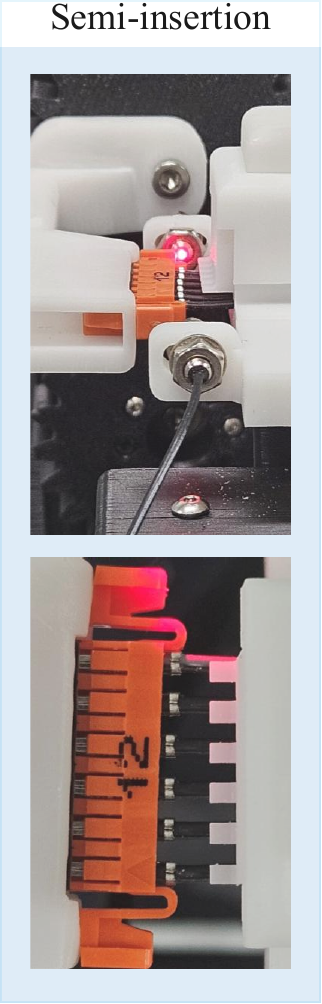}
    \label{Semi-insertion}
    }
    \hfill
    \subfigure[]{
    \centering
    \includegraphics[width=0.26\linewidth]{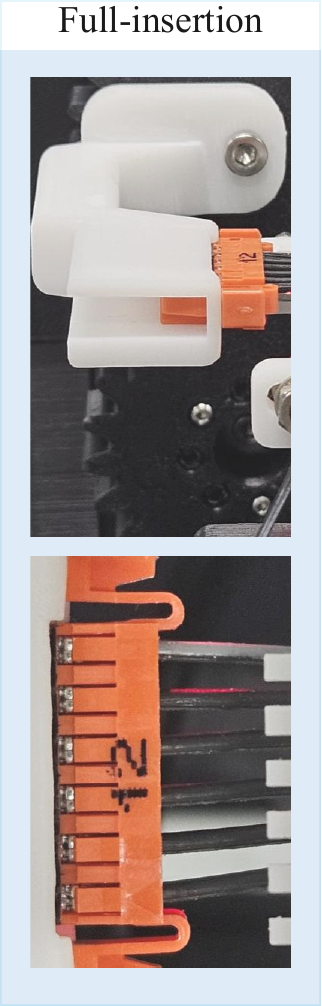}
    \label{Full-insertion}
    }
    \caption{Visual comparison of the insertion progress on both sides of
the connector. (a) Pre-insertion: terminal tips are aligned at the entrance of
the housing. (b) Semi-insertion: terminals are partially inserted with hooks
approaching the locking grooves. (c) Full-insertion: all terminals are fully
seated and locked into the housing.}
    \label{Insertion State}
\end{figure}

\subsection{Process 2 : Terminal Insertion Preparation}

\par Although terminals are aligned to the housing pitch at the guide outlet by Process~1, residual pose errors still hinder direct insertion into the housing. As noted in Sec.~\ref{subsec:2A}, roll and yaw errors are largely suppressed by the left–right partition walls of the guide; however, pitch error is relatively not corrected because no structure constrains the vertical direction, which leads to repeated insertion failures.

\par To minimize this uncertainty, the proposed harness assembly system adopts a three-stage strategy. First, \textit{Pre-insertion} (see Fig.~\ref{pre-insertion}) places each terminal tip partially on the corresponding housing hole to secure alignment. Second, \textit{Semi-insertion} (see Fig.~\ref{Semi-insertion}) pushes the terminals into the housing. Finally, \textit{Full-insertion} (see Fig.~\ref{Full-insertion}) completes the engagement. The following section describes the assembly mechanisms for each stage.

\begin{figure}[t]
    \centering
    \subfigure[]
    {
        \includegraphics[width=0.45\textwidth]{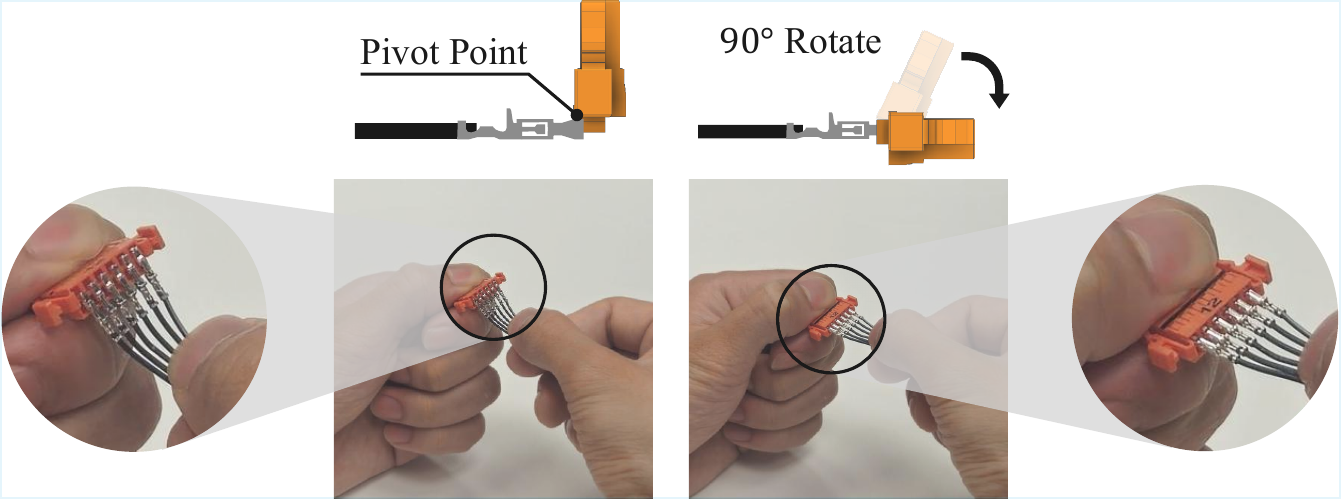}
        \label{fig:human_lns}
     }
    \hfill
    \subfigure[]{
        \includegraphics[width=0.48\textwidth]{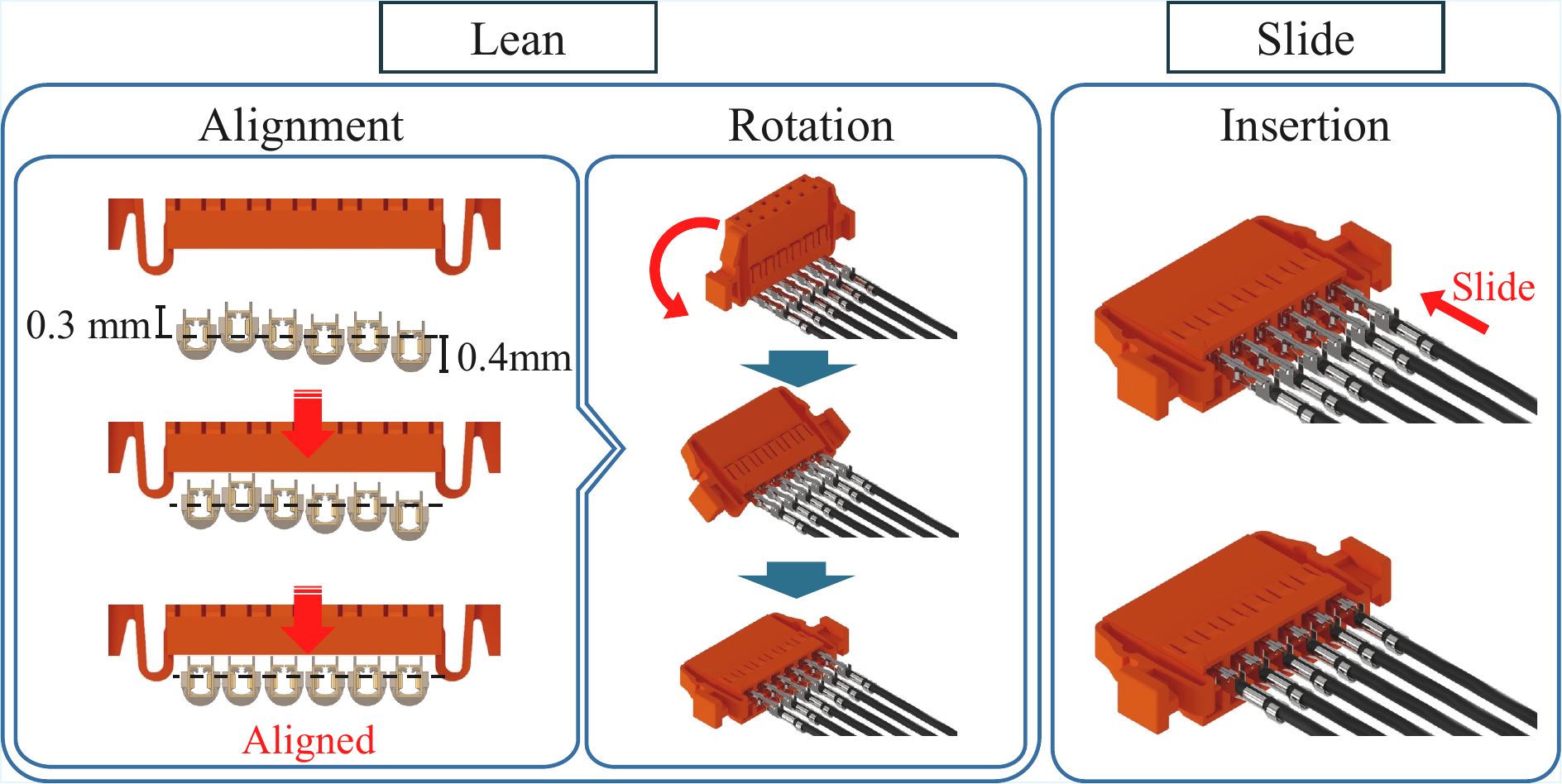}
        \label{fig:system_lns}
    }
    \caption{(a) Manual assembly operation of FRCH terminals that inspired the proposed Lean \& Slide technique. (b) Correction of pitch and yaw alignment errors using the Lean \& Slide method, which improves terminal alignment accuracy and enhances assembly reliability.}
    \label{fig:lean_slide_method}
\end{figure}

\subsubsection{Pre-insertion (Lean \& Slide)}
\label{LNS method}
\par We introduce a Lean \& Slide method that emulates manual assembly as shown in Fig.~\ref{fig:human_lns}. The housing first establishes a controlled contact at a hole edge and then rotates while sliding along the inlet surface so that the geometry passively aligns multiple terminals.
\par
\begin{figure}[t!]
    \centering
    \subfigure[]{
    \centering
    \includegraphics[width=0.21\linewidth]{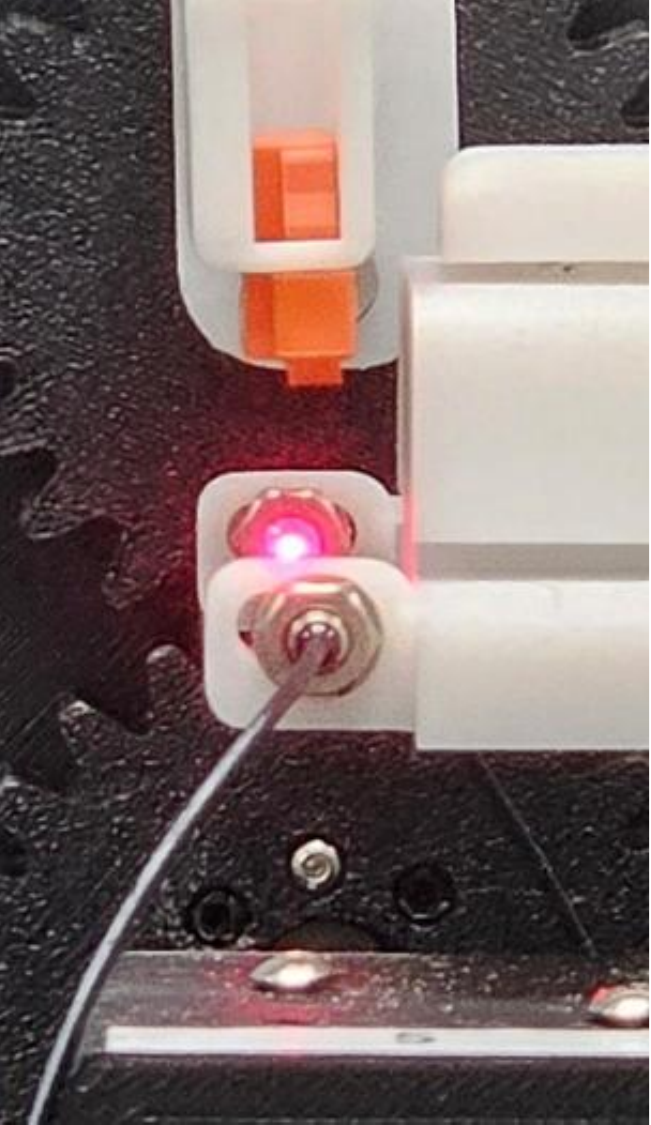}
    \label{fig:LnS_a}
    }
    \hfill
    \subfigure[]{
    \centering
    \includegraphics[width=0.21\linewidth]{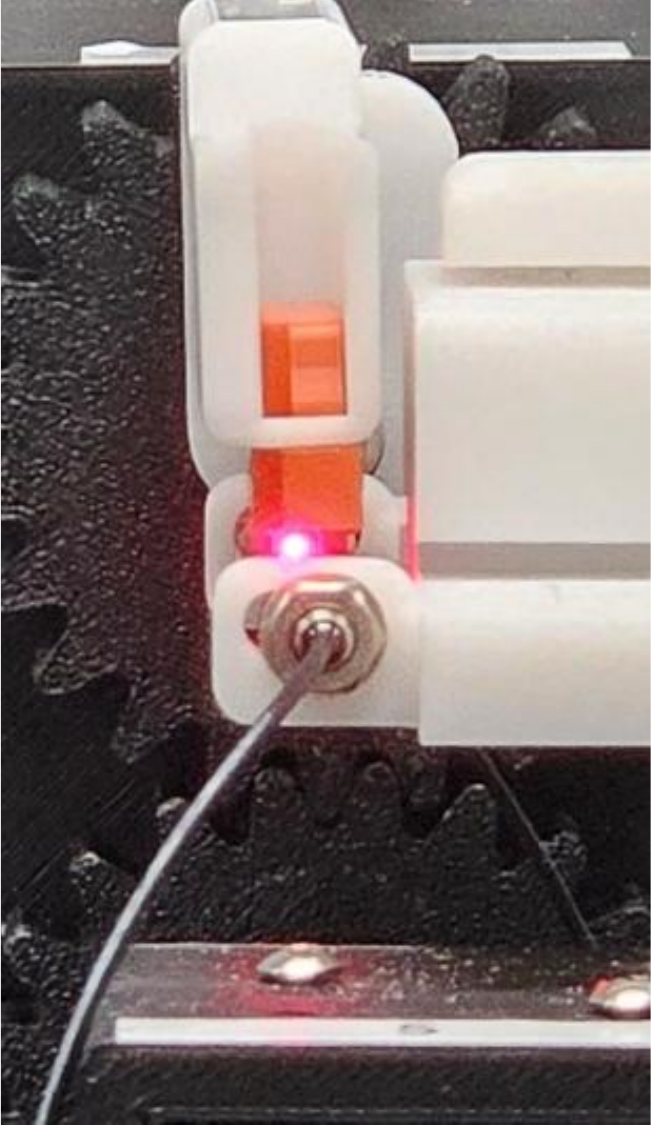}
    \label{fig:LnS_b}
    }
    \hfill
    \subfigure[]{
    \centering
    \includegraphics[width=0.21\linewidth]{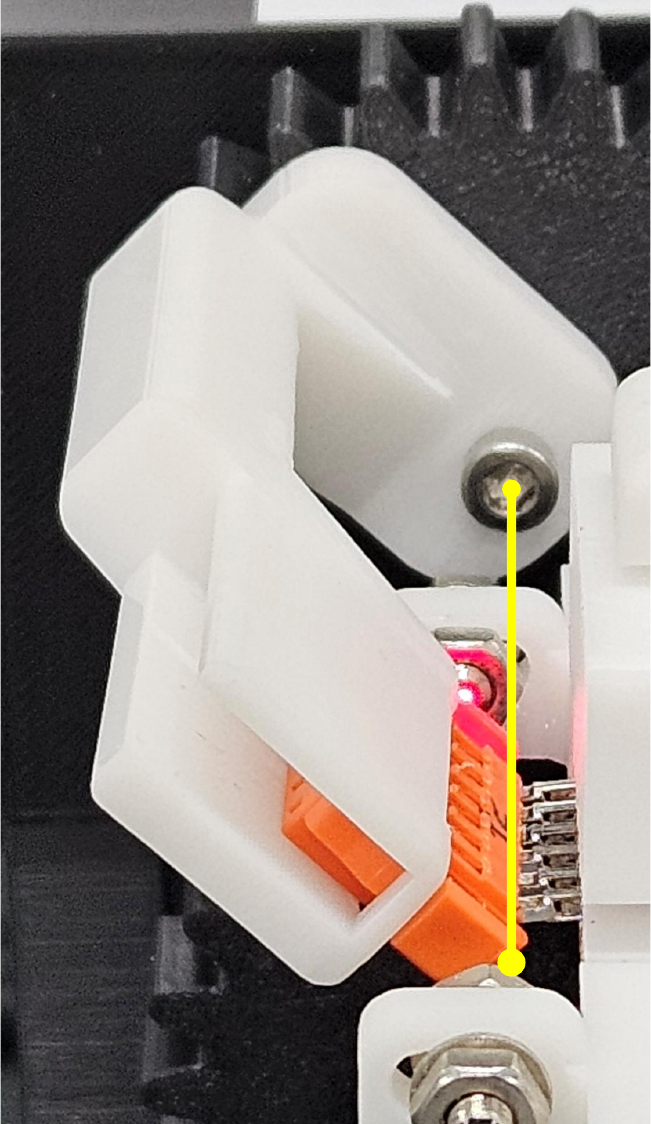}
    \label{fig:LnS_c}
    }
    \hfill
    \subfigure[]{
    \centering
    \includegraphics[width=0.21\linewidth]{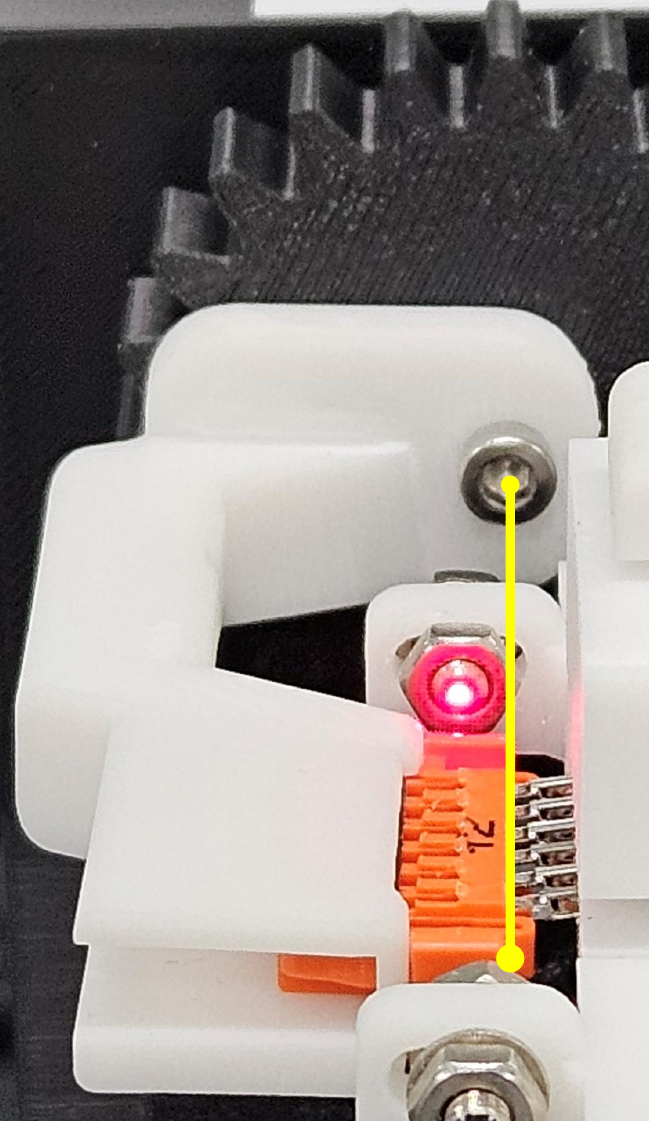}
    \label{fig:LnS_d}
    }
    \caption{Sequential motion of the rotor and terminal insertion process: (a) Initial position of the rotor, which moves along the Z-axis linear stage. (b) As the rotor descends along the Z-axis, the housing entrance sequentially contacts each terminal, compensating for pitch errors. (c) The rotor rotates around the pivot point (yellow line). (d) After the Lean \& Slide motion is completed, the terminal tips are inserted into the housing.}
    \label{fig:LnS_real}
\end{figure}

 As illustrated in Fig.~\ref{fig:system_lns}, the initial contact between the housing and the terminal faces provides a mechanical reference that cancels pitch error and aligns the terminal tips on a common plane. To establish this contact, the housing descends along the z-axis linear stage, as shown in Fig.~\ref{fig:LnS_real}\subref{fig:LnS_a}--\subref{fig:LnS_b}.
\par The housing then rotates about the contact pivot (see Fig.~\ref{fig:human_lns}), guiding each terminal into its corresponding hole. If a terminal grazes the housing surface during rotation, a slight roll motion is induced, allowing self-alignment to reduce residual pitch and yaw errors (see Fig.~\ref{fig:system_lns}).
\par A rotary–linear mechanism implements this procedure. Because Process~1 already sets inter-cable spacing, this stage focuses on a precise approach via rotation. The housing is coupled to the Rotor and initially positioned so that the terminal tips and housing holes overlap by half of \SI{0.9}{mm}, while maintaining vertical alignment to ensure smooth first contact. The rotation axis is mechanically coincident with the initial contact point, and continuous contact between the housing and terminals is maintained throughout \textit{Pre-insertion}. This single descend–contact–rotate sequence suppresses pitch and yaw errors without active control or expensive feedback, completing \textit{Pre-insertion} and enabling the subsequent \textit{Semi-insertion} stage.

\subsubsection{Semi-insertion (Weaving)}

\par
\begin{figure}[t]
    \centering
    \subfigure[]{
    \centering
    \includegraphics[width=0.45\linewidth]{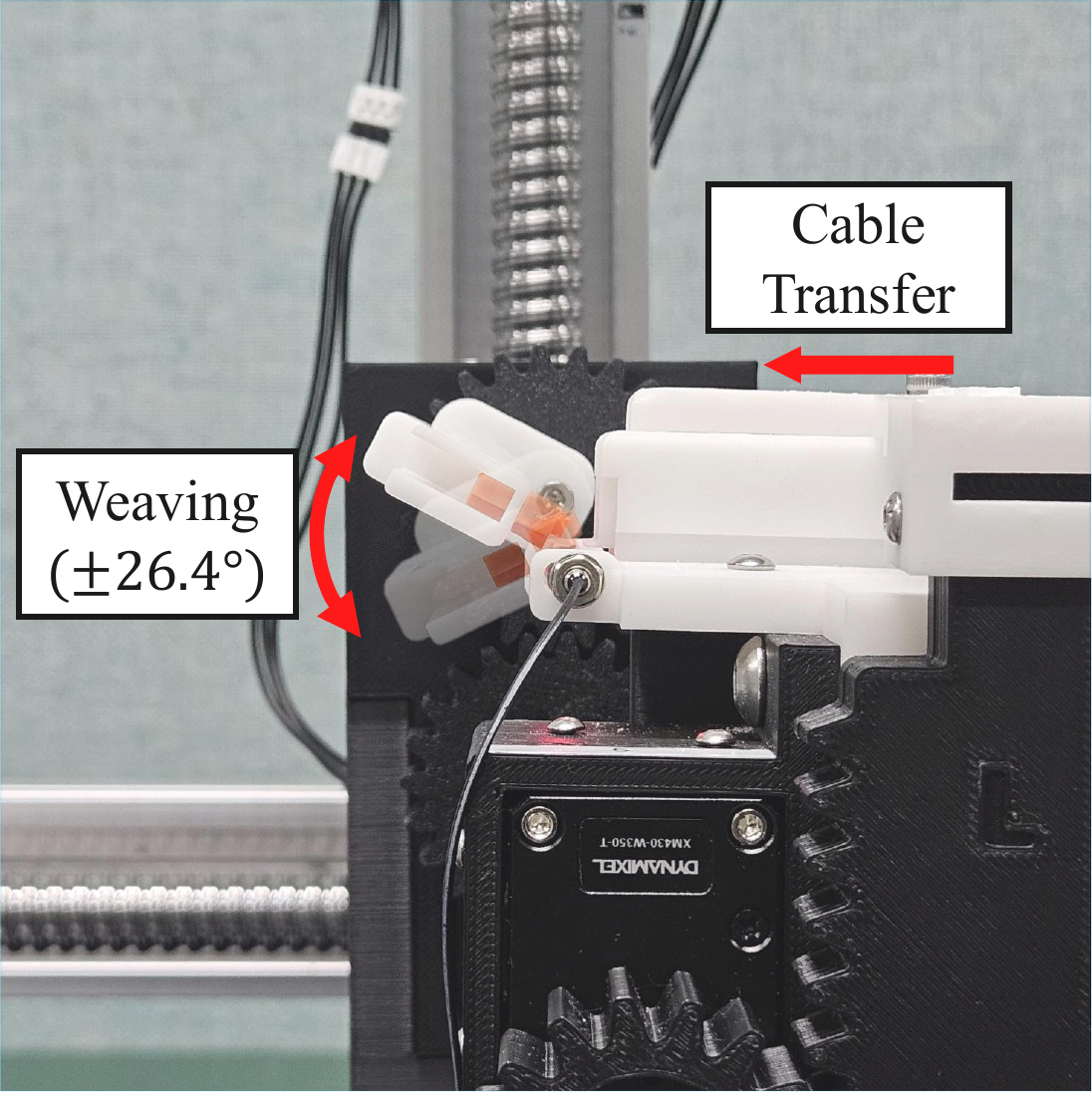}
    \label{fig:weave_left}
    }
    \hfill
    \subfigure[]{
    \centering
    \includegraphics[width=0.45\linewidth]{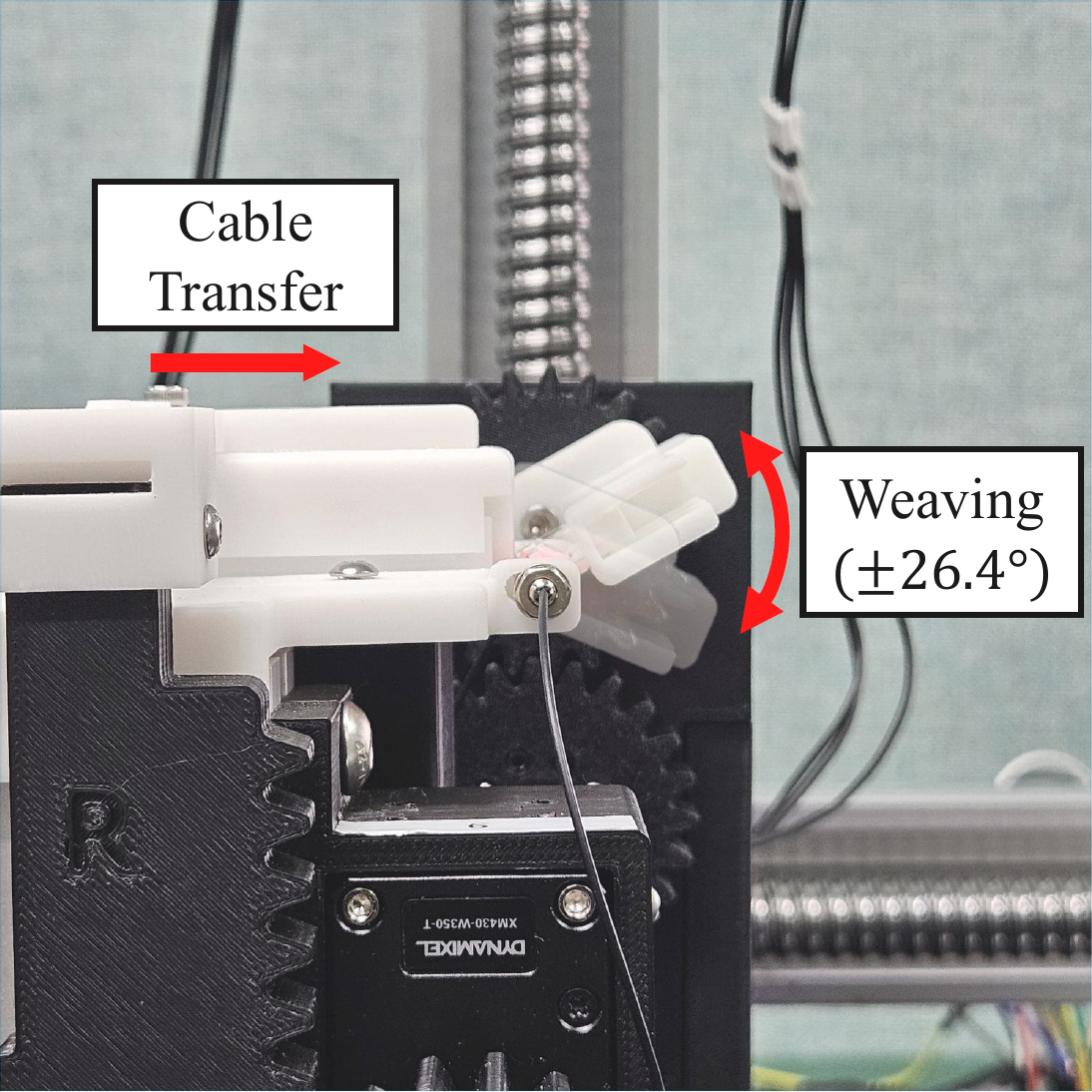}
    \label{fig:weave_right}
    }
    \caption{Weaving motion of the housing during semi-insertion: (a) left-side weaving with ±26.4° rotation, and (b) right-side weaving with ±26.4° rotation, while the cable is simultaneously transferred forward. }
    \label{fig:Weaving}
\end{figure}

 After the \textit{Pre-insertion}, the cable must be pushed further into the housing just before final locking. Due to the terminal latch geometry (see Fig.~\ref{fig:FRCH_fig}) and small protrusions inside the housing, internal catching can occur during insertion. We apply the weaving motion to relieve these interferences and promote insertion (see Fig.~\ref{fig:Weaving}).

\par In this stage, we move the housing relative to the stationary terminals to perform \textit{Semi-insertion}. As shown in Fig.~\ref{fig:Weaving}, the housing oscillates in the Pitch direction over a range of \(\pm 26.4^\circ\) at \(\SI{0.625}{Hz}\) for \(\SI{4}{s}\) while the Transfer Wheel simultaneously advances the cable at \(\SI{28.5}{mm/s}\) to promote insertion. The Weaving motion is executed simultaneously on the left and right housings. Upon completion, the terminals reach a state in which more than \(90\%\) of their length is inserted inside the housing. The latch is not yet engaged with the housing' locking window at this time; we define this condition as \textit{Semi-insertion} (see Fig.~\ref{Semi-insertion}).

\begin{figure}[t]
    \centering
    \subfigure[]{
    \centering
    \includegraphics[width=0.288\linewidth]{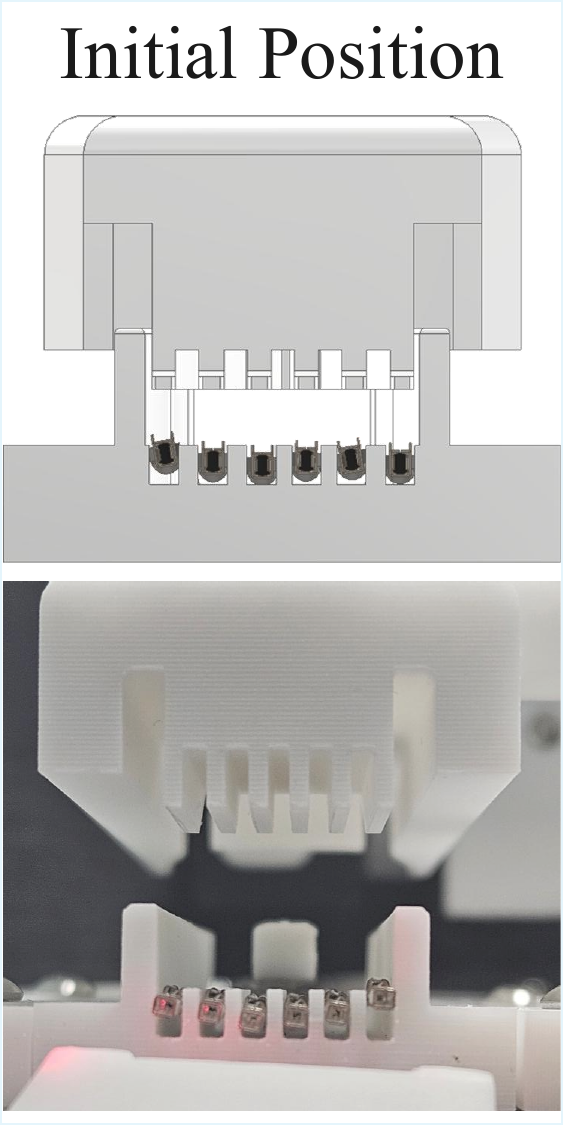}
    \label{fig:initial_position}
    }
    \hfill
    \subfigure[]{
    \centering
    \includegraphics[width=0.291\linewidth]{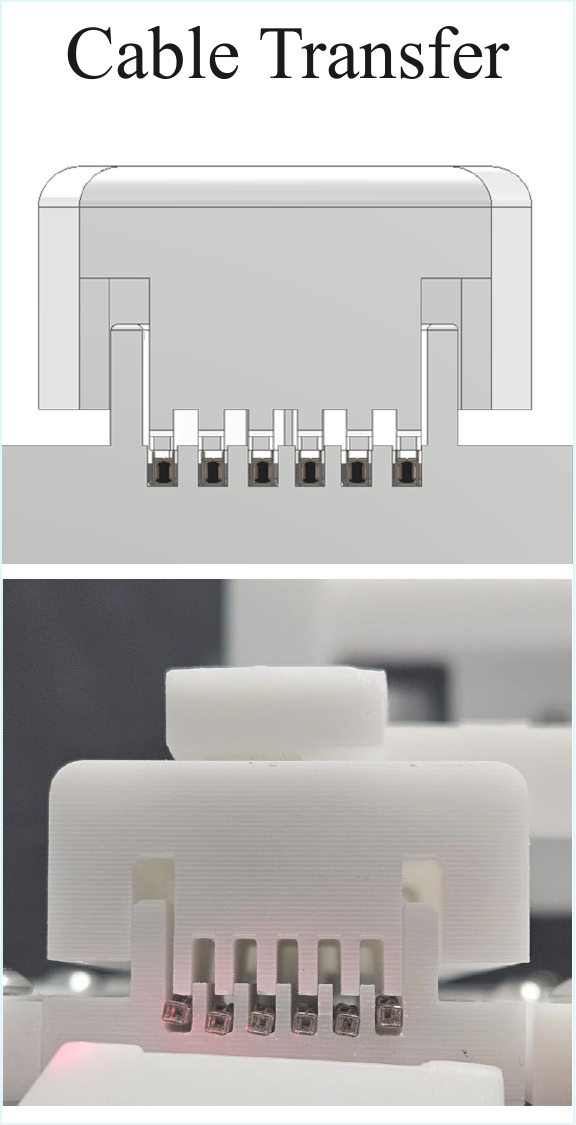}
    \label{fig:cable_transfer}
    }
    \hfill
    \subfigure[]{
    \centering
    \includegraphics[width=0.291\linewidth]{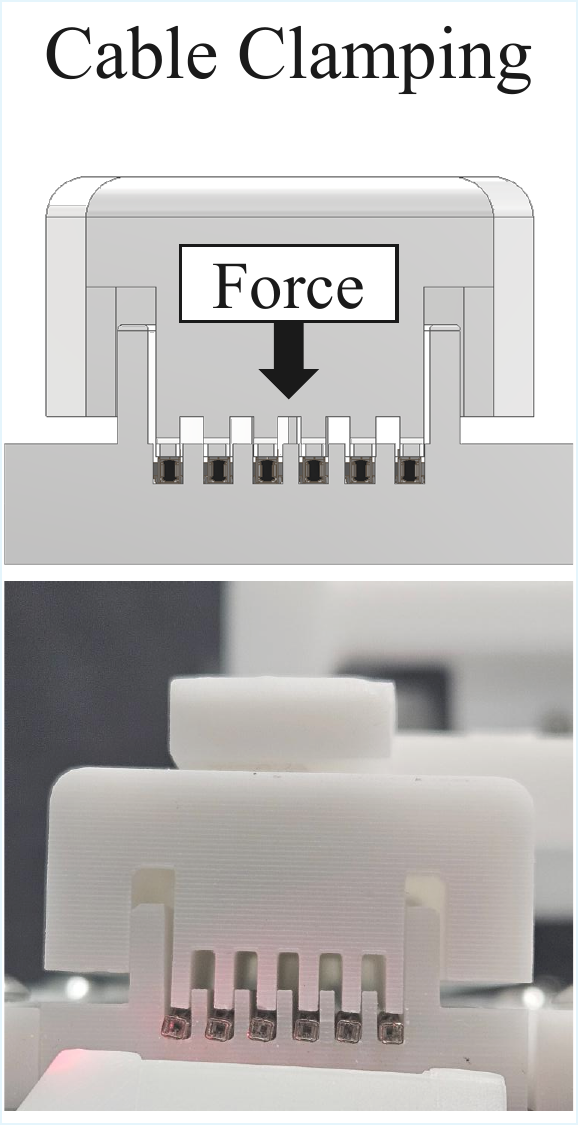}
    \label{fig:cable_clamping}
    }
    \caption{Cable positioning and clamping process using the guide top. (a) Initial position: a clearance is secured to allow cable retrieval to the outside. (b) Cable transfer: a clearance height of 2 mm is secured to allow cable movement. (c) Cable clamping: the guide top descends to a 1 mm height, applying downward force to eliminate position errors and fix the cable stably.}
    \label{fig:Guide_position}
\end{figure}

\begin{figure}[b]
    \centering
    \includegraphics[width=7.5cm]{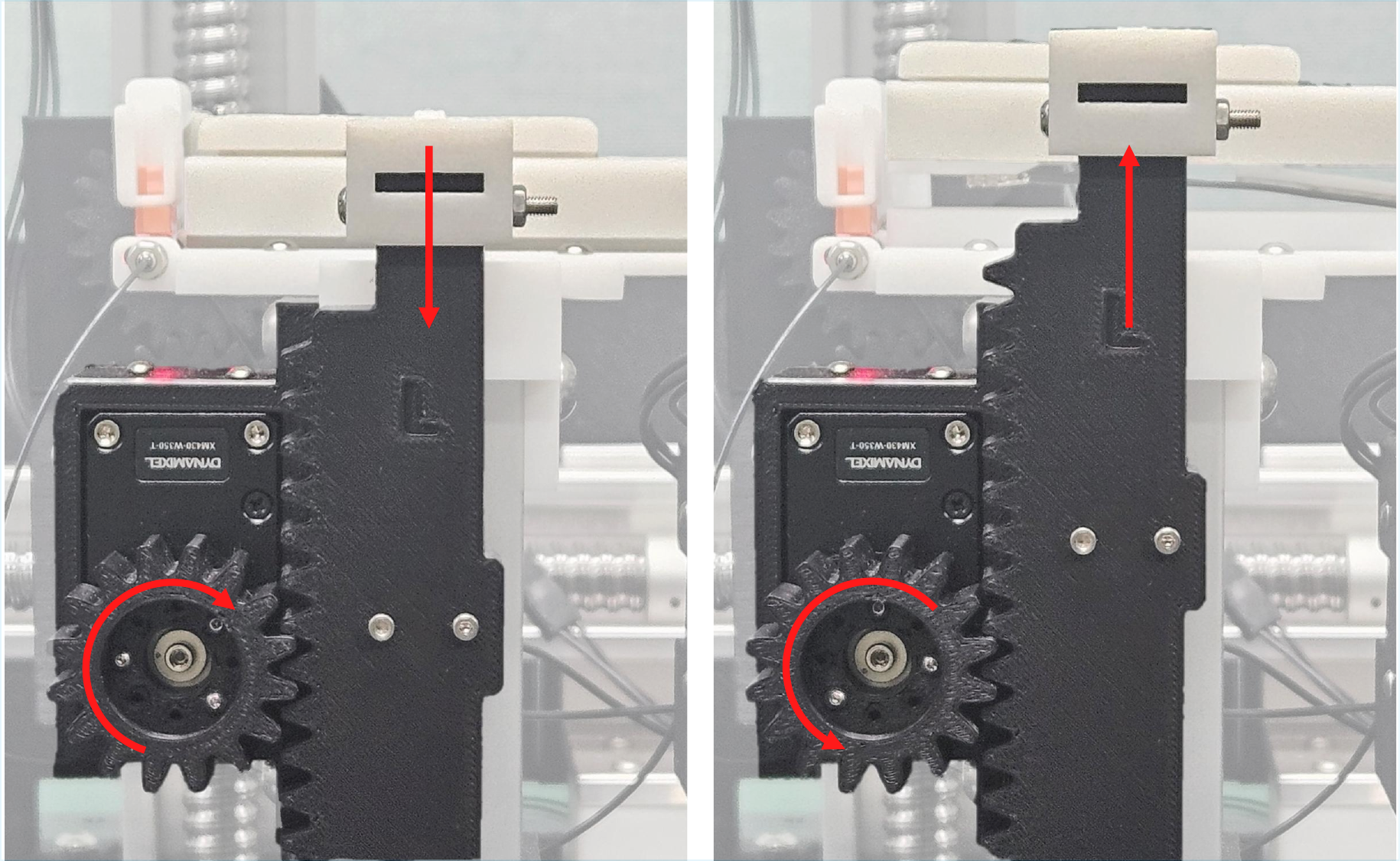}
    \caption{Vertical motion of the guide top using a rack-and-pinion mechanism. The servo motor rotates the pinion gear, which drives the rack to move the guide top upward and downward for cable clamping.}
    \label{fig:guide_top_clamping}
    \vspace{-0.2cm}
\end{figure}

\begin{figure*}[t]
    \centering
    \includegraphics[width=0.8\textwidth]{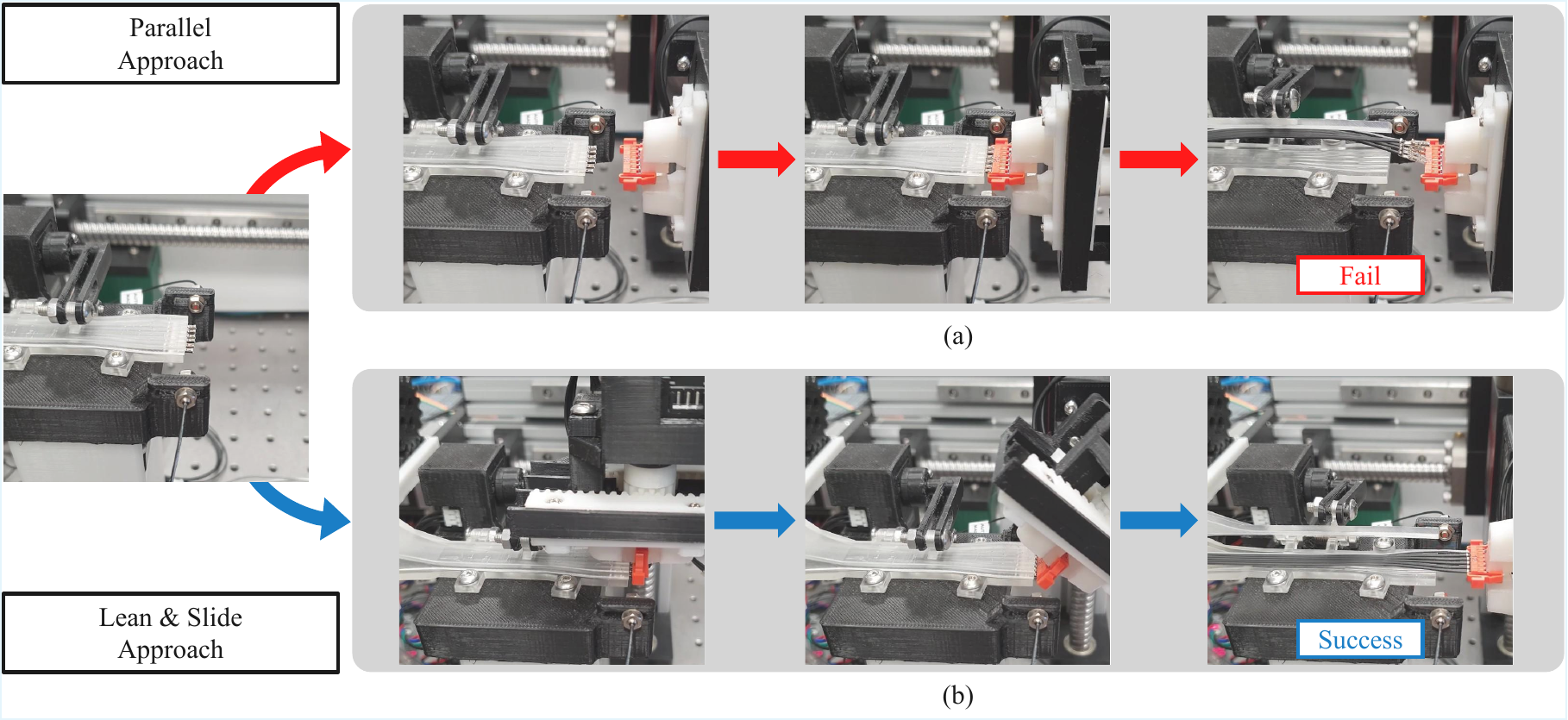}
    \caption{Comparison of insertion approaches.
 (a) Parallel approach: housing is aligned horizontally without correction, resulting in misalignment and insertion failure.
 (b) Lean \& Slide approach: housing is tilted and rotated during insertion, enabling self-alignment of terminals and achieving successful insertion.}
    \label{fig:parallel vs lns}
    \vspace{-0.2cm}
\end{figure*}

\subsection{Process 3 : Full Insertion (Clamping)}

\par To compensate for position errors and improve insertion reliability in harness-cable assembly, we designed an integrated structure that incorporates a clamping function into the guide top (see Fig.~\ref{fig:Aligner Assembled}). As described in Section~\ref{process1_Guide_transfer}, the guide bottom and top form a semi-closed pair that engages to open and close while allowing height adjustment. This structure not only aligns the cables but also prevents terminal detachment and wobbling during insertion, and it provides structural constraints that ensure the alignment state is maintained through the Full-insertion stage.

\par During the Cable Transfer stage, the guide top clamps at a height that allows the cable to pass freely (see Fig.~\ref{fig:cable_transfer}). At this time, the guide top functions not as a terminal fixture but as a lid of the guide path, enabling smooth forward motion of the cable. After Semi-insertion is completed, the guide top descends and interlocks with the guide bottom to perform a clamping action that secures the cable (see Fig.~\ref{fig:cable_clamping}). The downward motion of the guide top is actuated by a Rack \& Pinion mechanism (see Fig.~\ref{fig:guide_top_clamping}). The slot spacing is modularized to match the terminal pitch, which ensures maintainability by allowing replacement or servicing at the level of individual slots when necessary.

\par Clamping prevents micro-vibration and backward slip that may occur during the Full-insertion stage, thereby improving stability. It also suppresses cross-interference among cable bundles with multiple degrees of freedom and preserves alignment, which enhances assembly reliability at Full insertion. The mechanism remains effective in high-speed, repetitive insertion environments.

\begin{figure}[htbp]
    \centering
    \includegraphics[width=8.0cm]{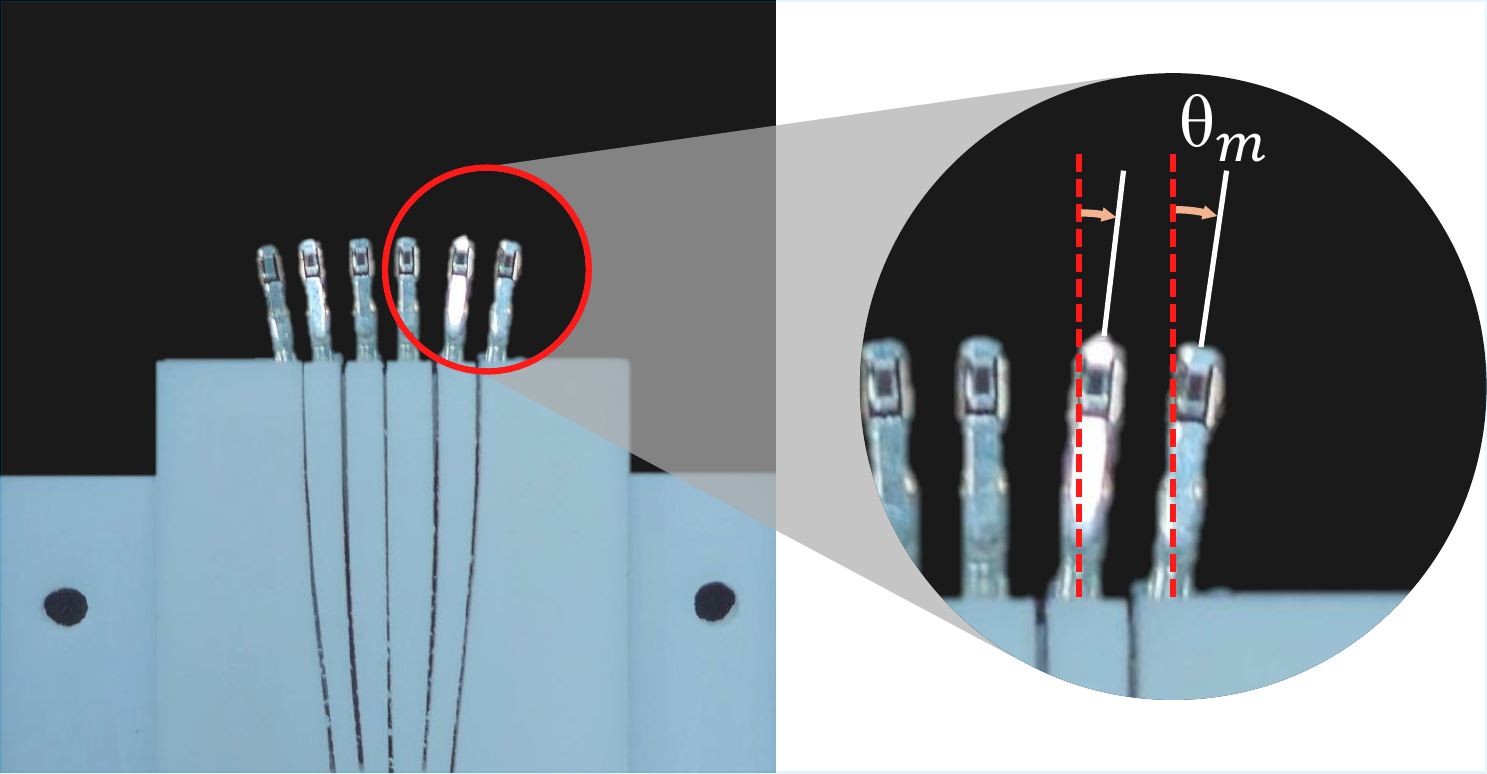}
     \caption{Residual yaw error observed at the terminals after passing through the guide. The terminal misalignment angle $\theta_m$ remains even after spacing adjustment, which prevents successful parallel insertion and highlights the necessity of a new correction mechanism. }
    \label{fig: curved error}
    \vspace{-0.2cm}
\end{figure}

\section{Assembly Experiment} \label{sec:exp}

\subsection{Comparison of Parallel and Lean \& Slide Approaches}

\begin{figure*}[htbp]
    \centering
    \includegraphics[width=0.8\textwidth]{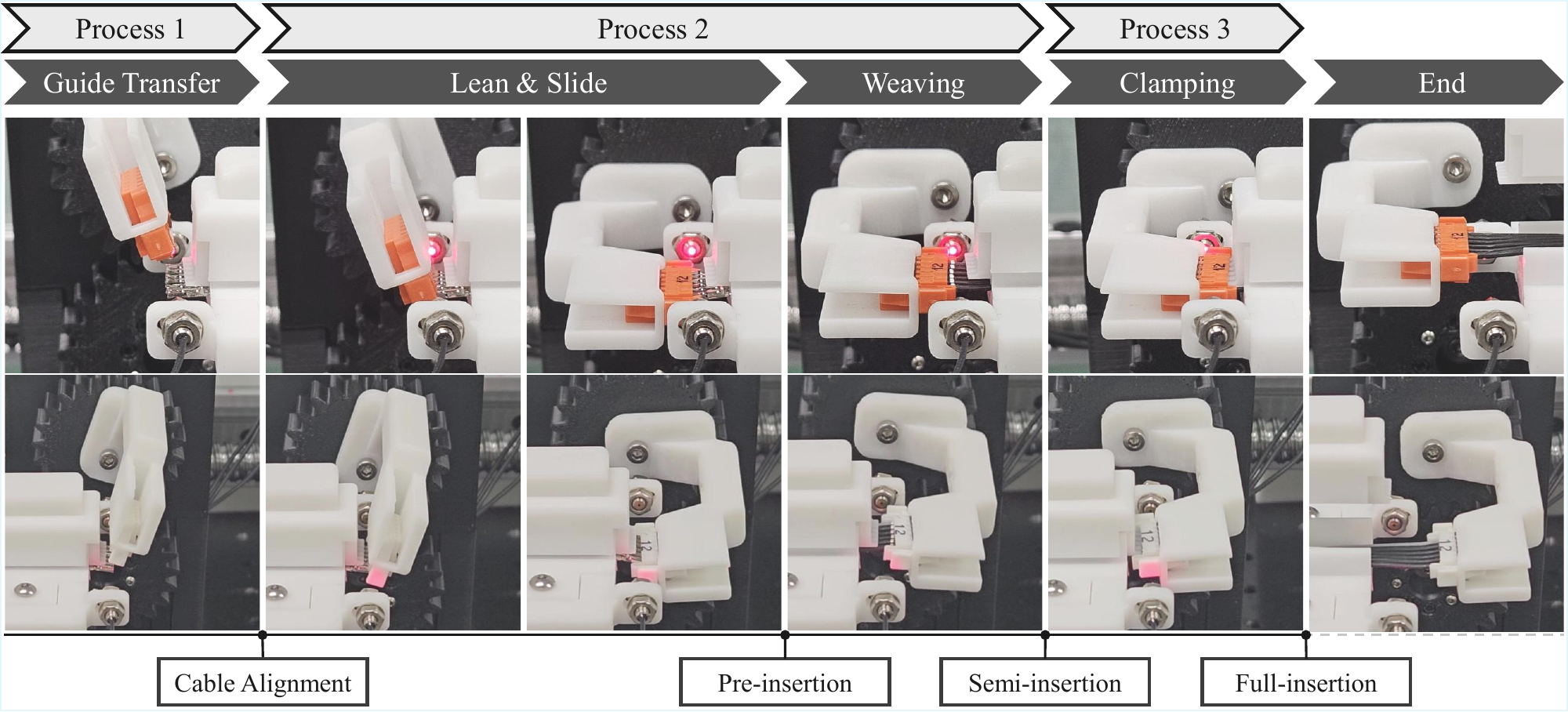}
    \caption{Integrated workflow of the proposed experimental procedure for cable terminal insertion. The process consists of three main stages: (1) cable alignment, (2) terminal insertion preparation (pre-insertion and semi-insertion), and (3) full-insertion. These stages are realized through five consecutive steps: (1) guide transfer, (2) lean \& slide, (3) weaving, (4) clamping, and (5) end.}
    \label{fig: Procedure Overview}
    \vspace{-0.2cm}
\end{figure*}

\par Before conducting the full assembly experiment, we performed a study to verify the effect of the Lean \& Slide mechanism. For cables aligned by the guide, we compared a method in which the housing approaches horizontally and inserts (see Fig.~\ref{fig:parallel vs lns} (a), Parallel Approach) with a method in which the housing first approaches vertically and then rotates by \(90^\circ\) to assemble as introduced in Section~\ref{LNS method} (see Fig.~\ref{fig:parallel vs lns} (b), Lean \& Slide Approach). A trial was regarded as a failure if any one of the six terminals was not inserted.

\par The Parallel Approach achieved a success rate of \(15\%\), as summarized in TABLE ~\ref{tab:approach experiment}. The primary failure mode was the inability to remove terminal orientation error, with failures due to pitch and yaw errors being particularly prominent. After the cable bundle passes through the guide, a residual terminal misalignment may remain at the terminal tips. In this study, the terminal misalignment angle $\theta_m$ is definded as the yaw-direction angular deviation between the longitudinal axis of each terminal and the intended housing insertion direction. As shown in Fig.~\ref{fig: curved error}, multiple cables of identical length tended to form an arced configuration after guide transfer. 
Because the guide primarily regulates inter-terminal spacing, it has limited capability to eliminate terminal orientation errors. In particular, yaw and pitch errors exceeded nominal tolerances by up to \(5^\circ\), which frequently caused terminals to catch and fail to enter the housing apertures.

\begin{table}[b]
\caption{Success Rate Comparison of Approach Methods }
\centering
\resizebox{0.95\linewidth}{!}{
\renewcommand{\arraystretch}{1.2}
\begin{tabular}{ccccl}
\toprule\hline
Method&Success& \multicolumn{3}{c}{Failure*}\\
\cline{3-5}
&& Roll error&Pitch error&Yaw error\\
\hline
Parallel & 3 / 20 (15\%)& 1 / 20& 10 / 20&6 / 20\\

Lean \& Slide& 18 / 20 (90\%)& 0 / 20& 0 / 20&2 / 20\\
\end{tabular}}\\[2mm]
{\footnotesize \parbox{0.9\linewidth}{*Roll, Pitch, and Yaw errors indicate terminal orientation errors exceeding the allowable range relative to the housing hole direction.}}
\label{tab:approach experiment}
\end{table}

\par In contrast, the Lean \& Slide Approach achieved a success rate of \(90\%\), demonstrating a substantial improvement over the Parallel Approach. This result shows that Lean \& Slide is particularly effective in preventing failures caused by pitch error, which was the primary failure mode in the Parallel Approach. During Lean \& Slide, the housing first establishes controlled contact with the terminal tips. It then rotates while sliding along the inlet surface. This motion passively corrects residual pitch misalignment and guides the terminals into the corresponding housing apertures. As a result, stable \textit{Pre-insertion} was achieved in most trials. Some failures still occurred when the initial cable pose was excessively twisted or deformed. In such cases, the input pose exceeded the correction range of the passive alignment mechanism.

\subsection{Entire Experiment Procedure}

\par We conducted the full assembly experiment, including the proposed Lean \& Slide mechanism, and the complete assembly sequence covering Processes 1--3 is shown in Fig.~\ref{fig: Procedure Overview}. A trial was regarded as successful if, after assembly, the cable was correctly locked into the housing with complete engagement and no visible defects. Conversely, a trial was classified as a failure in cases of incomplete cable insertion, a misassembled cable order, or the housing was not fully engaged, such that external defects were observed.

\begin{table}[b]
\caption{End-to-End and Phase-wise Success Rates}
\centering
\resizebox{0.95\linewidth}{!}{
\renewcommand{\arraystretch}{1.3}
\begin{tabular}{cccc}
\toprule\hline
\multicolumn{2}{c}{Metric}& Success&95\% CI\\
\hline
\multicolumn{2}{c}{Total Process}& 67 / 80 (83.75\%)&[74.16\%, 90.25\%]\\

 & └ First 40 trials& 34 / 40 (85.0\%)&\\

 & └ Last 40 trials& 33 / 40 (82.5\%)&\\
\hline
 \multicolumn{2}{c}{Transfer phase
}& 43 / 50 (86.0\%)&[73.81\%, 93.05\%]
\\
 \multicolumn{2}{c}{Insertion phase}& 49 / 50 (98.0\%)&[89.5\%, 99.65\%]
\end{tabular}}
\label{tab:experiment success rate}
\end{table}

\par A total of 80 end-to-end assembly trials were conducted to evaluate the overall performance of the proposed system. Among these trials, 67 were successful, resulting in an end-to-end process success rate of \(83.75\%\), as summarized in TABLE ~\ref{tab:experiment success rate}. To ensure prototype safety, the servo motor output was limited to approximately half of its maximum level, and under this condition, the total assembly process took approximately \SI{33}{s}. When the output was set to its maximum level, the cycle time was reduced to about \SI{15}{s}. To examine robustness under repeated operation, the 80 end-to-end trials were further divided into two equal halves of 40 trials each. In the first half, 34 out of 40 trials were successful, corresponding to a success rate of \(85.0\%\), whereas in the second half, 33 out of 40 trials were successful, corresponding to a success rate of \(82.5\%\). Although the success rate in the second half was slightly lower than that in the first half, the difference was marginal and does not indicate a substantial performance drop. These results suggest that the proposed mechanism maintained stable repeatability over repeated operation without a meaningful degradation trend.

\par To analyze the causes of process failures, checkpoints were defined throughout the assembly procedure. The overall process can be divided into two major stages: the cable alignment stage before coupling with the housing (Process 1) and the actual insertion stage where the housing engages and assembly is completed (Processes 2–3). These two stages are referred to as the Transfer phase and the Insertion phase, respectively. By dividing the process in this manner and establishing checkpoints for each phase, it was possible to distinguish the preparation procedure from the insertion procedure and to systematically analyze their respective success rates and failure modes. Separate standalone experiments were then conducted for the two phases. In the Transfer phase, 43 out of 50 trials were successful, yielding a success rate of \(86.0\%\). In the Insertion phase, which was evaluated under successful cable alignment conditions, 49 out of 50 trials were successful, corresponding to a success rate of \(98.0\%\). These phase-wise results indicate that failures occurred more frequently during the cable transfer and alignment stage than during the insertion stage after successful alignment. The results of the end-to-end trials, repeatability analysis, and phase-wise experiments are summarized in TABLE ~\ref{tab:experiment success rate}.

\section{Discussion and Conclusion} \label{sec:dis and con}

\par We successfully implemented one-row, bidirectional FRCH assembly for two different housing types, achieving a total cycle time of 33~s that is shorter than prior peg-in-hole–based studies. The overall process success rate reached \(83.75\%\). In phase-wise experiments, the Transfer phase achieved an \(86.0\%\) success rate, and the Insertion phase achieved a \(98.0\%\) success rate. These results indicate that failures occurred more frequently during the \textit{Cable Alignment} step, i.e., the Transfer phase. We analyze that the cable’s high degrees of freedom introduce diverse disturbances while the cable is transferred through the guide, and that control uncertainty increases as the distance between the transfer wheel applying force to the cable and the terminal grows during transport.

\par Although these metrics do not fully meet industrial automation requirements, this study demonstrates, at laboratory scale, the feasibility of automated terminal-to-housing assembly for FRCH, a process for which no prior automated example has been reported to the best of our knowledge. Unlike conventional automated systems that rely on unidirectional assembly, the proposed system implements bidirectional simultaneous assembly tailored to the characteristics of the complex FRCH. Beyond the specific application, this work formulates FRCH terminal-to-housing assembly as a mechanically coupled multi-terminal insertion problem and demonstrates a sensor-minimal mechanical strategy for addressing correlated alignment errors, insertion interference, and unstable intermediate states.

\par The proposed approach also has potential scalability and adaptability because the assembly process is composed of modular mechanical operations: Guide Transfer, Lean \& Slide, Weaving, and Clamping. For single-row FRCH products, the same assembly principle can be adapted by modifying product-dependent mechanical parameters, such as terminal count, terminal pitch, guide-path curvature, housing holder geometry, Lean \& Slide contact position, Weaving conditions, and clamping conditions. For longer single-row FRCHs, additional guide channels and intermediate cable-support structures may be introduced to reduce accumulated pose uncertainty. For multi-row housings, the same \textit{Cable Alignment}–\textit{Pre-insertion}–\textit{Semi-insertion}–\textit{Full-insertion} principle may be extended through row-wise insertion or vertically stacked/indexed guide and clamping modules.

\par The current system has several limitations. Although the proposed assembly principle can be adapted to other FRCH products, changes in terminal count, terminal pitch, cable stiffness, or housing geometry require reconfiguration of the guide path, housing holder, pivot point, weaving condition, and clamping condition. This product-specific setup may increase the process preparation time, especially when the system is applied to a new cable or connector family. In addition, increasing the number of terminals or rows may amplify correlated misalignment, cable compliance, and insertion interference. Therefore, further hardware and control extensions are required to support different housing geometries, longer terminal arrays, and multi-row assembly. In addition, the current hardware is a 3D-printed prototype, which limits dimensional accuracy, mechanical stiffness, durability, and cycle-time performance under industrial conditions. 

\par Future work will focus on improving the \textit{Cable Alignment} step, where failures occurred relatively often. In particular, we plan to apply a gripper that transports the cable while grasping the stiffer crimped terminal section, thereby reducing variability during transfer. We also plan to incorporate vision-based failure detection and retry motions to improve process reliability. Finally, precision mechanisms manufactured from high-strength materials and industrial servo drives will be introduced to improve speed, durability, and long-term repeatability. 
\par Overall, this study provides a laboratory-scale proof of feasibility for automating FRCH terminal-to-housing assembly, which remains a challenging task in harness-cable assembly. To the best of our knowledge, no prior work has demonstrated simultaneous alignment and insertion of multiple mechanically coupled terminals in a ribbon-cable harness environment. The proposed approach can serve as a basis for future automation of flexible, densely arranged, and mechanically coupled assembly tasks.

\raggedbottom

\ifCLASSOPTIONcaptionsoff
  \newpage
\fi


%

\bibliographystyle{IEEEtran}
\bibliography{References}

@article{karlsson2024automatic,
  author  = {Karlsson, T. and {\AA}blad, E. and Hermansson, T. and Carlson, J. S. and Tenf{\"a}lt, G.},
  title   = {Automatic Cable Harness Layout Routing in a Customizable 3D Environment},
  journal = {Computer-Aided Design},
  volume  = {169},
  pages   = {103671},
  year    = {2024},
  doi     = {10.1016/j.cad.2023.103671}
}

@article{hernandezmejia2026review,
  author  = {Hernandez-Mejia, M. and Romero, D. and Ruppert, T. and
             Guedea, F. and Salunkhe, O. and Rodriguez, C. A. and
             Stahre, J.},
  title   = {A Review of Robotic Manipulation Solutions for Deformable
             Linear Objects: The Case of Wire Harnesses (Co-)Assembly
             by Robots},
  journal = {Robotics and Autonomous Systems},
  volume  = {199},
  pages   = {105375},
  year    = {2026},
  doi     = {10.1016/j.robot.2026.105375}
}

@article{navasreascos2026simulation,
  author  = {Navas-Reascos, Gabriel E. and Romero, David and
             Guedea, Federico and Rodriguez, Ciro A. and Stahre, Johan},
  title   = {Simulation Tests of Wire Harness Assembly Tasks Supported
             by Collaborative Robots for Different Types of Wire Harnesses},
  journal = {International Journal of Computer Integrated Manufacturing},
  year    = {2026},
  doi     = {10.1080/0951192X.2026.2619775},
  note    = {In press}
}

@article{lorenz2023approaches,
  author  = {Lorenz, N. and Mayer, R.},
  title   = {Approaches for Automated Wiring Harness Manufacturing:
             Function Integration with Additive Manufacturing},
  journal = {Automotive and Engine Technology},
  volume  = {8},
  number  = {4},
  pages   = {227--237},
  year    = {2023},
  doi     = {10.1007/s41104-023-00137-9}
}

@article{NGUYEN2024360,
  title   = {Revolutionizing robotized assembly for wire harness: A {3D} vision-based method for multiple wire-branch detection},
  journal = {Journal of Manufacturing Systems},
  volume  = {72},
  pages   = {360--372},
  year    = {2024},
  issn    = {0278-6125},
  doi     = {10.1016/j.jmsy.2023.12.002},
  author  = {Thong Phi Nguyen and Donghyung Kim and Hyun-Kyo Lim and Jonghun Yoon}
}

@article{9268291,
  author={Yumbla, Francisco and Abeyabas, Meseret and Luong, Tuan and Yi, June-Sup and Moon, Hyungpil},
  booktitle={2020 20th International Conference on Control, Automation and Systems (ICCAS)}, 
  title={Preliminary Connector Recognition System Based on Image Processing for Wire Harness Assembly Tasks}, 
  year={2020},
  volume={},
  number={},
  pages={1146-1150},
  doi={10.23919/ICCAS50221.2020.9268291}}

@article{ying2022pose,
  title={Pose estimation of a small connector attached to the tip of a cable sticking out of a circuit board},
  author={Ying, Changjian and Mo, Yaqiang and Matsuura, Yuichiro and Yamazaki, Kimitoshi},
  journal={International Journal of Automation Technology},
  volume={16},
  number={2},
  pages={208--217},
  year={2022},
  publisher={Fuji Technology Press Ltd.}
}

@ARTICLE{10746551,
  author={Mei, Yecheng and Li, Ruiya and Tan, Yuegang and Zhu, Ding and Li, Tianliang and Zhou, Zude},
  journal={IEEE Transactions on Automation Science and Engineering}, 
  title={Robotic Assembly Strategy With Wrist Force Sense for Narrow Clearance Peg-in-Hole}, 
  year={2025},
  volume={22},
  number={},
  pages={8709-8724},
  doi={10.1109/TASE.2024.3487988}}

@ARTICLE{9064951,
  author={Xing, Dengpeng and Liu, Xiwei and Liu, Fangfang and Xu, De},
  journal={IEEE Transactions on Industrial Informatics}, 
  title={Efficient Insertion Strategy for Precision Assembly With Uncertainties Using a Passive Mechanism}, 
  year={2021},
  volume={17},
  number={2},
  pages={1263-1273},
  doi={10.1109/TII.2020.2986805}}

@ARTICLE{8365152,
  author={Zhang, Kuangen and Xu, Jing and Chen, Heping and Zhao, Jianguo and Chen, Ken},
  journal={IEEE Transactions on Industrial Electronics}, 
  title={Jamming Analysis and Force Control for Flexible Dual Peg-in-Hole Assembly}, 
  year={2019},
  volume={66},
  number={3},
  pages={1930-1939},
  doi={10.1109/TIE.2018.2838069}}

@ARTICLE{9748081,
  author={Xu, Jingjing and Liu, Kang and Pei, Yanhu and Yang, Congbin and Cheng, Yanhong and Liu, Zhifeng},
  journal={IEEE Transactions on Instrumentation and Measurement}, 
  title={A Noncontact Control Strategy for Circular Peg-in-Hole Assembly Guided by the 6-DOF Robot Based on Hybrid Vision}, 
  year={2022},
  volume={71},
  number={},
  pages={1-15},
  doi={10.1109/TIM.2022.3164133}}

@article{NGUYEN2021365,
  title    = {A novel vision-based method for {3D} profile extraction of wire harness in robotized assembly process},
  journal  = {Journal of Manufacturing Systems},
  volume   = {61},
  pages    = {365--374},
  year     = {2021},
  issn     = {0278-6125},
  doi      = {10.1016/j.jmsy.2021.10.003},
  author   = {Thong Phi Nguyen and Jonghun Yoon}
}

@article{MOU2022105164,
  title    = {Pose estimation and robotic insertion tasks based on {YOLO} and layout features},
  journal  = {Engineering Applications of Artificial Intelligence},
  volume   = {114},
  pages    = {105164},
  year     = {2022},
  issn     = {0952-1976},
  doi      = {10.1016/j.engappai.2022.105164},
  author   = {Fangli Mou and Hao Ren and Bin Wang and Dan Wu}
}

@ARTICLE{9650911,
  author={Chen, Zhong and Xie, Shengyang and Zhang, Xianmin},
  journal={IEEE Transactions on Instrumentation and Measurement}, 
  title={Position/Force Visual-Sensing-Based Robotic Sheet-Like Peg-in-Hole Assembly}, 
  year={2022},
  volume={71},
  number={},
  pages={1-11},
  doi={10.1109/TIM.2021.3135552}}

@ARTICLE{8395267,
  author={De Gregorio, Daniele and Zanella, Riccardo and Palli, Gianluca and Pirozzi, Salvatore and Melchiorri, Claudio},
  journal={IEEE Transactions on Automation Science and Engineering}, 
  title={Integration of Robotic Vision and Tactile Sensing for Wire-Terminal Insertion Tasks}, 
  year={2019},
  volume={16},
  number={2},
  pages={585-598},
  doi={10.1109/TASE.2018.2847222}}

@ARTICLE{10220113,
  author={Men, Yu and Jin, Ligang and Cui, Tao and Bai, Yunfeng and Li, Fengming and Song, Rui},
  journal={IEEE Transactions on Instrumentation and Measurement}, 
  title={Policy Fusion Transfer: The Knowledge Transfer for Different Robot Peg-in-Hole Insertion Assemblies}, 
  year={2023},
  volume={72},
  number={},
  pages={1-10},
  doi={10.1109/TIM.2023.3305709}}

@ARTICLE{10721256,
  author={Hao, Tiantian and Xu, De},
  journal={IEEE Transactions on Industrial Informatics}, 
  title={Automated Control Method of Multiple Peg-in-Hole Assembly for Relay and Its Socket}, 
  year={2025},
  volume={21},
  number={1},
  pages={990-998},
  doi={10.1109/TII.2024.3476535}}

@article{beck2025deep,
  author  = {Beck, L. and Gebauer, D. and Rauh, T. and others},
  title   = {Deep learning-based localization of electrical connector sockets for automated mating},
  journal = {Production Engineering Research and Development},
  volume  = {19},
  pages   = {187--194},
  year    = {2025},
  doi     = {10.1007/s11740-024-01299-7}
}

@article{Caporali2026VisionTactile,
  author  = {Alessio Caporali and Michele Mirto and Salvatore Pirozzi and Gianluca Palli},
  title   = {Vision and Tactile Sensing for {DLO} Manipulation and Pin Insertion in Robotic Connector Assembly},
  journal = {IEEE/ASME Transactions on Mechatronics},
  year    = {2026},
  note    = {Early Access},
  doi     = {10.1109/TMECH.2026.3654272}
}

@article{Wang2024ComputerVisionWireHarness,
  author  = {Hao Wang and Omkar Salunkhe and Walter Quadrini and Dan L{\"a}mkull and Fredrik Ore and M{\'e}lanie Despeisse and Luca Fumagalli and Johan Stahre and Bj{\"o}rn Johansson},
  title   = {A Systematic Literature Review of Computer Vision Applications in Robotized Wire Harness Assembly},
  journal = {Advanced Engineering Informatics},
  volume  = {62},
  pages   = {102596},
  year    = {2024},
  doi     = {10.1016/j.aei.2024.102596}
}

@article{Cho2025CableWiring,
  author  = {Youngsu Cho and Minsu Cho and Jongwoo Park and Byung-Kil Han and Young Hun Lee and Sung-Hyuk Song and Chanhun Park and Dong Il Park},
  title   = {Strategic Algorithm for Cable Wiring Using Dual Arm with Compliance Control},
  journal = {Robotics and Computer-Integrated Manufacturing},
  volume  = {93},
  pages   = {102924},
  year    = {2025},
  doi     = {10.1016/j.rcim.2024.102924}
}

@article{Hartisch2024ConnectorAssembly,
  author  = {Richard M. Hartisch and Kevin Haninger},
  title   = {High-Speed Electrical Connector Assembly by Structured Compliance in a {Finray-Effect} Gripper},
  journal = {IEEE/ASME Transactions on Mechatronics},
  volume  = {29},
  number  = {2},
  pages   = {810--819},
  year    = {2024},
  doi     = {10.1109/TMECH.2023.3324227}
}

\end{document}